\documentclass{article}
\usepackage{soul,color}

    \PassOptionsToPackage{numbers, compress}{natbib}
    \usepackage[preprint]{neurips_2024}

\usepackage[utf8]{inputenc} % allow utf-8 input
\usepackage[T1]{fontenc}    % use 8-bit T1 fonts
\usepackage{hyperref}       % hyperlinks
\usepackage{url}            % simple URL typesetting
\usepackage{booktabs}       % professional-quality tables
\usepackage{amsfonts}       % blackboard math symbols
\usepackage{nicefrac}       % compact symbols for 1/2, etc.
\usepackage{microtype}      % microtypography
\usepackage{xcolor}         % colors

\usepackage{makecell}

\usepackage{booktabs}
\usepackage{multirow}
\usepackage{algorithm}
\usepackage{algorithmic}
\usepackage{graphicx}
\usepackage{amsmath}
\usepackage{amssymb}
\usepackage{mathtools}
\usepackage{amsthm}
\usepackage{enumitem}
\usepackage{lscape}
\usepackage{derivative}

\usepackage[capitalize,noabbrev]{cleveref}

\newcommand{\E}{\mathbb{E}}
\newcommand{\R}{\mathbb{R}}
\newcommand{\boldx}{\boldsymbol{x}}

\newcommand{\boldV}{\boldsymbol{V}}
\newcommand{\btheta}{\boldsymbol{\theta}}
\newcommand{\bLambda}{\boldsymbol{\Lambda}}
\newcommand{\calC}{\mathcal{C}}
\newcommand{\calD}{\mathcal{D}}
\newcommand{\calN}{\mathcal{N}}

\title{Enhancing Generative Molecular Design via Uncertainty-guided Fine-tuning of Variational Autoencoders}

\author{%
  A N M Nafiz Abeer \\
  Department of Electrical and Computer Engineering\\
  Texas A\&M University\\
  College Station, TX, USA \\
  \texttt{nafiz.abeer@tamu.edu} \\
  \And
   Sanket Jantre \\
  Computational Science Initiative\\
  Brookhaven National Laboratory\\
  Upton, NY, USA \\
  \texttt{sjantre@bnl.gov} \\
  \And 
  Nathan M Urban \\
  Computational Science Initiative\\
  Brookhaven National Laboratory\\
  Upton, NY, USA \\
  \texttt{nurban@bnl.gov } \\
  \And
   Byung-Jun Yoon \\
  Department of Electrical and Computer Engineering\\
  Texas A\&M University\\
  College Station, TX, USA \\
  Computational Science Initiative\\
  Brookhaven National Laboratory\\
  Upton, NY, USA \\
  \texttt{bjyoon@tamu.edu} \\
}

\begin{document}

\maketitle

\begin{abstract}
In recent years, deep generative models have been successfully adopted for various molecular design tasks, particularly in the life and material sciences. A critical challenge for pre-trained generative molecular design (GMD) models is to fine-tune them to be better suited for downstream design tasks aimed at optimizing specific molecular properties. However, redesigning and training an existing effective generative model from scratch for each new design task is impractical. Furthermore, the black-box nature of typical downstream tasks--such as property prediction--makes it nontrivial to optimize the generative model in a task-specific manner. In this work, we propose a novel approach for a model uncertainty-guided fine-tuning of a pre-trained variational autoencoder (VAE)-based GMD model through performance feedback in an active learning setting. 
% The main idea is to learn a low-dimensional active subspace of the high-dimensional VAE parameters to efficiently capture model uncertainty via Bayesian inference, given the training data. 
The main idea is to quantify model uncertainty in the generative model, which is made efficient by working within a low-dimensional active subspace of the high-dimensional VAE parameters explaining most of the variability in the model’s output. The inclusion of model uncertainty expands the space of viable molecules through decoder diversity. We then explore the resulting model uncertainty class via black-box optimization made tractable by low-dimensionality of the active subspace. This enables us to identify and leverage a diverse set of high-performing models to generate enhanced molecules. Empirical results across six target molecular properties, using multiple VAE-based generative models, demonstrate that our uncertainty-guided fine-tuning approach consistently outperforms the original pre-trained models.
\end{abstract}

\section{Introduction}
\label{Sec:Introduction}
Machine learning has evolved significantly in the field of drug discovery, with early focus on quantitative structure-activity relationship (QSAR) \cite{QSAR} for high-throughput screening (HTS) \cite{clyde2021high, woo2023optimal}, is now attracting research interest in de-novo molecule design, driven by the rise of deep generative models. Such models \cite{gomez2018automatic, PSO_cont_latent_space, jtvae_paper, hiergraph} allow exploration of molecular space using optimization algorithms in a low-dimensional latent space derived from high-dimensional chemical data. However, the effectiveness of these generative models in creating target molecules is constrained by their training datasets, as is the case with any data-driven approach. Depending on downstream task--such as generating valid molecules with optimum properties using specific reactants--research efforts have focused on either optimization algorithm \cite{olivecrona2017, constrained_BO, chance-constrained, decoder_uncertainty} with minimal model changes or completely redesign the generative model \cite{ORGANIC, li2018multi, conditional_gen, jin20b_multi, mars}. Improving performance for new downstream tasks often requires rethinking generative model design. However, finding a universally effective design remains challenging, as evidenced by the aforementioned works. By building upon existing pre-trained models, we aim to leverage the unique insights embedded within those models from the collective experience of the research community to enhance their performance in various downstream design tasks of interest.

Fine-tuning a generative model for a quantity of interest in a molecular design task can be challenging, especially with limited data \cite{blaschke2021finetune}. We address this challenge by efficiently quantifying the model uncertainty by employing active subspace reduction of a generative model \cite{abeer2024leveraging}, which constructs a low-dimensional subspace of the generative model parameters capturing most of the variability in the model's output. Incorporating model uncertainty leads to diversity in VAE model parameters (specifically the decoder in our problem setting) which expands the space of viable molecules compared to the pre-trained model. First, we assume that optimization over the latent space of a pre-trained variational autoencoder (VAE) model yields a list of candidate designs. These candidates, which can result from multiple runs of some optimization procedure with different hyperparameters, are decoded to generate molecules that determine the model's downstream performance. For these candidates within latent space, we adapt the generative model in its low-dimensional active subspace to enhance its performance beyond that of the pre-trained model. We achieve this through black-box optimization, guided by performance feedback from downstream tasks. This optimization tunes the distribution of active subspace parameters to generate diverse models that outperform the pre-trained model for those candidate latent points.
% Black-box optimization, guided by downstream task feedback, tunes the distribution of active subspace parameters to generate diverse models surpassing the pre-trained model's performance. 
The black-box nature of our optimization for improving model performance in downstream molecular design tasks simplifies its integration with existing optimization methods in latent space of VAE-based generative models.

To this end, our contributions are as follows:
% \vspace{-1mm}
\begin{itemize}[topsep=0pt,itemsep=1pt,parsep=2pt,partopsep=0pt]
    \item We explore the model uncertainty class of VAE-based generative models, effectively represented by the low-dimensional active subspace parameters, using black-box optimization algorithms: Bayesian optimization (BO) and REINFORCE. The proposed fine-tuning approach leads to diverse high-performing models which then improve the generative model's performance in downstream design tasks of interest. 
    \item We empirically demonstrate the effectiveness of our uncertainty-guided fine-tuning approach in leveraging model uncertainty to enhance downstream performance across six target molecular properties. Our method consistently improves the design performance of multiple pre-trained VAE-based models through the proposed closed-loop optimization scheme.
\end{itemize}

\section{Problem Setting}
\subsection{VAE-based Generative Models for Molecular Design}
\label{GMD_VAE}
Across different VAE-based generative models for molecular design tasks \cite{gao2022sample}, the encoder component of a VAE generally embeds molecular representations into a low-dimensional continuous latent space, while the decoder converts the latent space embeddings back to the chemical space. For inverse molecular design, various optimization approaches can be applied on the learned latent space of a trained VAE. For example, \cite{gomez2018automatic} trained a VAE jointly with a property predictor network using SMILES representation of molecules. Subsequently, a Gaussian process (GP) was trained to predict target properties from the latent representation, leading to some latent points corresponding to high-scoring molecules. Similarly, \cite{jtvae_paper} used Bayesian optimization on the latent space of their Junction Tree VAE (JT-VAE) model to generate molecules with optimized properties. 

\begin{figure*}[tb]
\vskip 0.2in
\begin{center}
\centerline{\includegraphics[width=0.9\textwidth]{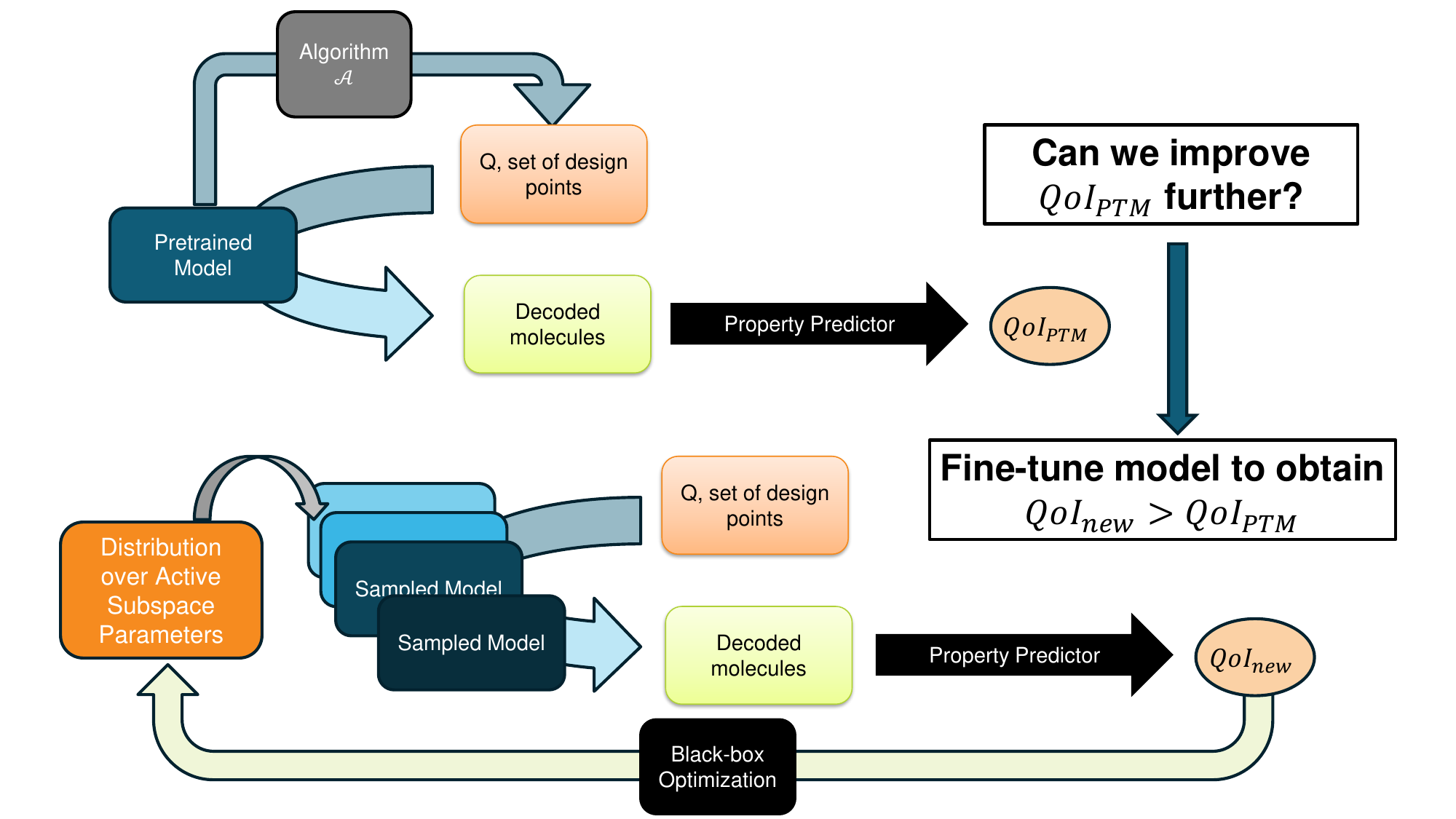}}
\caption{Illustration of the quantity of interest ($QoI$) enhancement problem. Using a \underline{p}re-\underline{t}rained VAE-based generative \underline{m}odel (PTM), an algorithm $\mathcal{A}$ finds a set of design points -- $Q$ in its latent space. As a downstream task, a property predictor is applied to the molecules corresponding to $Q$ to obtain the pre-trained model' $QoI$ ($QoI_{\text{PTM}}$). Our objective is to fine-tune the model parameters to further enhance $QoI$ for the same $Q$. We propose to leverage the active subspace of model parameters and perform black-box optimization over the subspace parameters with $QoI$ feedback.}
\label{framework_fig}
\end{center}
\vskip -0.2in
\end{figure*}

\subsection{Downstream Performance of Pre-trained Models}
Given a pre-trained VAE model, we want to use it for a downstream task of generating molecules with desired properties. Let's denote the pre-trained model by $\mathcal{M}_{\boldsymbol{\theta}_0}$ where $\boldsymbol{\theta}_0$ are the pre-trained model parameters. For the downstream task of interest -- $\mathcal{T}$, algorithm $\mathcal{A}$ (e.g. Bayesian optimization in the work of \cite{gomez2018automatic,jtvae_paper}) is applied in conjunction with the pre-trained model $\mathcal{M}_{\boldsymbol{\theta}_0}$ to look for candidate points within the latent space of  $\mathcal{M}_{\boldsymbol{\theta}_0}$ so that properties of the molecules corresponding to those candidates are optimized.
% For example, $\mathcal{A}$ can search within the latent space based on the properties of molecules decoded by  $\mathcal{M}_{\boldsymbol{\theta}_0}$.
% For example, $\mathcal{A}$ can be Bayesian optimization in the work of \cite{gomez2018automatic,jtvae_paper} mentioned in \cref{GMD_VAE}.
Specifically, for a given pre-trained model (PTM), the algorithm $\mathcal{A}$ finds a set of candidate design points, $Q$, which $\mathcal{M}_{\boldsymbol{\theta}_0}$ decodes to generate corresponding molecules. The properties of these molecules define the quantity of interest ($QoI$) of the pre-trained model $QoI_{\text{PTM}}$, e.g. average property value of top $10\%$ molecules.

Our contention is that while the set $Q$ may achieve the best $QoI$ for the pre-trained model, the algorithm $\mathcal{A}$ can perform better if the VAE model is fine-tuned for task -- $\mathcal{T}$. However, fine-tuned models are not always available for the task at hand. In this work, we investigate whether we can tune a given pre-trained model so that the molecules generated from the set $Q$ (found by $\mathcal{A}$ using $\mathcal{M}_{\boldsymbol{\theta}_0}$) achieve a $QoI$ better than $QoI_{\text{PTM}}$. Here, the set $Q$ contains the candidate latent points found by some optimization procedure in the latent space of pre-trained VAE, and $QoI_{\text{PTM}}$ is some target property statistics over the associated molecules. Our goal is to bias the pre-trained model to produce molecules with better $QoI$ for the same design points in $Q$.

\subsection{Problem Definition}
To summarize our objective, as illustrated in \cref{framework_fig}, we assume a set of design points -- $Q$ has been found using an algorithm $\mathcal{A}$ applied to the pre-trained VAE model. These design points can be decoded by the model to reconstruct molecules. We aim to further optimize the model parameters to generate better molecules from $Q$ than the pre-trained model does. Denoting the $QoI$ of the pre-trained model as $QoI_{\text{PTM}} = \phi(\mathcal{M}_{\boldsymbol{\theta}_0},Q)$, our goal is
\begin{align}
    \max_{\boldsymbol{\theta} \in \boldsymbol{\Theta}} \phi(\mathcal{M}_{\boldsymbol{\theta}},Q)
    \label{opt_problem}
\end{align}
If the quantity of interest, $\phi$, can be predicted with a separate predictor network using the generative model's output, then gradient-based fine-tuning can approximately solve (\ref{opt_problem}). However, the downstream task can be complex, making the $QoI$ or $QoI$-related proxy prediction difficult, and requiring separate predictors for different tasks. Alternatively, treating $\phi$ as a black-box function is computationally challenging due to the high dimensionality of the model's parameter space, $\boldsymbol{\Theta}$, making black-box optimization methods like Bayesian optimization difficult to apply. \cite{abeer2024leveraging} demonstrated that models sampled within the low dimensional active subspace of JT-VAE are diverse enough to affect the molecule generation from the latent space of the pre-trained model. In this work, we utilize the active subspace within the VAE model parameter space $\boldsymbol{\Theta}$ as design space for black-box optimization.

\section{Method}
\label{Sec:Method}
\subsection{Active Subspace within Neural Network Parameter Space}
\label{AS_summary}
The active subspace (AS) of deep neural networks, as described by \cite{jantre2024active}, aims to identify a low-dimensional subspace in the high-dimensional neural network parameter space that have the most influence on the network's output. Given a neural network $f_{\btheta}(\boldx)$ with input -- $\boldx$ and stochastic network parameters -- $\btheta \in \R^D$ following probability distribution $p(\btheta)$, we can construct an uncentered covariance matrix of the gradients: $\calC = \E_{\btheta}\left[(\nabla_{\btheta} f_{\btheta}(\boldx)) (\nabla_{\btheta} f_{\btheta}(\boldx))^T\right]$. If $\calC$ admits the eigendecomposition: $\calC = \boldV \bLambda \boldV^T$ where $\boldV$ includes the eigenvectors and $\bLambda=\rm{diag}(\lambda_1, \dots, \lambda_D)$ are the eigenvalues with $\lambda_1 \ge \dots \lambda_D \ge 0$. We then can extract $k$ dimensional active subspace by partitioning $\boldV$ into $[\boldV_1, \boldV_2]$ where $\boldV_1 \in \R^{D\times k}$ and $\boldV_2 \in \R^{D \times (D-k)}$ with $k \le n \ll D$.  Accordingly, the active subspace is spanned by $\boldV_1$ corresponding to the largest $k$ eigenvalues.
% In case of a pre-trained model with parameters $\btheta_0$, the expectation is taken according to a probability distribution centered around $\btheta_0$, e.g. $\mathcal{N}(\btheta_0, \sigma_0^2 \mathbf{I})$ where $\sigma_0$ is perturbation noise standard deviation. For further details of the active subspace of deep neural networks, readers are referred to \cite{jantre2024active}.

\subsection{Optimization over Active Subspace}
\label{opt_section}
Similar to \cite{abeer2024leveraging}, we have considered two disjoint partitions of the parameter space $\boldsymbol{\Theta}$: $\boldsymbol{\Theta}^S$ -- containing set of stochastic parameters, $\boldsymbol{\theta}^S$, and $\boldsymbol{\Theta}^D$ -- containing rest of the deterministic parameters, $\boldsymbol{\theta}^D$. In \cref{intuition_tree_decoder}, we discuss how we can intuitively distribute the generative model's components into these two partitions for solving the optimization problem in \cref{opt_problem}. Instead of directly approximating the epistemic uncertainty of the stochastic parameters $\boldsymbol{\theta}^S$, we construct the active subspace $\boldsymbol{\Omega}$ within $\boldsymbol{\Theta}^S$ while keeping the parameters in $\boldsymbol{\Theta}^D$ fixed at their pre-trained values, i.e. $\boldsymbol{\theta}^D = \boldsymbol{\theta}_0^D$. Specifically, we learn the projection matrix, $\mathbf{P}$, which maps the active subspace parameters, $\boldsymbol{\omega} \in \boldsymbol{\Omega}$, to its corresponding parameter space, $\boldsymbol{\Theta}^S$, as follows
\begin{equation}
\btheta^S = \btheta_0^S + \mathbf{P} \boldsymbol{\omega}
\label{proj_eqn}   
\end{equation}
To construct the active subspace, we compute the gradient (only for the parameters in $\boldsymbol{\Theta}^S$) of the loss function that is used to train the generative model $\mathcal{M}_{\boldsymbol{\theta}_0}$. In this work, we considered the combination of reconstruction loss and the KL divergence loss of the VAE models as the $f_{\btheta}(\boldx)$ mentioned in \cref{AS_summary} while freezing the parameters in $\boldsymbol{\Theta}^D$ to $\boldsymbol{\theta}_0^D$. Next, we apply the variational inference method \cite{Blei2017} to approximate the posterior distribution of $\boldsymbol{\omega}$, i.e. $p(\boldsymbol{\omega} \vert \calD)$ using the training dataset of the pre-trained model. Specifically, we learn the active subspace posterior distribution parameters by minimizing the sum of training loss of VAE model and the KL divergence loss between approximated posterior distribution and the prior distribution over the active subspace parameters. Details of constructing the active subspace and posterior approximation are given in \cref{appendix: AS}.

During the inference stage, we draw $M$ samples, $\left\{ \boldsymbol{\omega}_i\right\}_{i=1}^{M}$ independently from approximated $p(\boldsymbol{\omega} \vert \calD)$, and using (\ref{proj_eqn}) we have $M$ number of model instances in parameter space $\boldsymbol{\Theta}$. Hence for the downstream task, we now have a diverse pool of models instead of a single pre-trained model. To quantify the uncertainty of the model's output, we can perform the Bayesian model averaging with this collection of models as in \cite{abeer2024leveraging}. Next, we use the distribution over active subspace parameters $\boldsymbol{\omega}$ as the design space for finding a collection of models suitable for producing molecules with better $QoI$ over set $Q$.

Under the variational inference, we learn the approximate posterior distribution over active subspace parameters as an uncorrelated multivariate normal distribution parameterized by $\boldsymbol{\mu}_{\text{post}}$ and $\boldsymbol{\sigma}_{\text{post}}$. 
Our optimization goal is to fine-tune these distribution parameters to improve pre-trained model's $QoI$ on the fixed design set $Q$. Denoting the fine-tuned distribution parameters as $\boldsymbol{\mu}_{f}$ and $\boldsymbol{\sigma}_{f}$, we can rewrite the optimization problem in (\ref{opt_problem}) as follows
\begin{equation}
\max_{\boldsymbol{\mu}_f,\boldsymbol{\sigma}_{f}} \phi(\boldsymbol{\mu}_f,\boldsymbol{\sigma}_{f},Q)
\label{red_opt_problem}    
\end{equation}
The $QoI$ function $\phi$ in (\ref{red_opt_problem}) is evaluated using models sampled from an active subspace parameter distribution with corresponding distribution parameters being $\boldsymbol{\mu}_f, \boldsymbol{\sigma}_{f}$. $M$ independent samples are drawn from $p(\boldsymbol{\omega} ; \boldsymbol{\mu}_f,\boldsymbol{\sigma}_{f})$ which lead to a collection of models, $\{\mathcal{M}_{\boldsymbol{\theta}_i}\}_{i=1}^{M}$, using (\ref{proj_eqn}). The design points in $Q$ are uniformly distributed among these $M$ models for decoding. The property of interest is predicted for the reconstructed molecules and the predicted values are summarized e.g. average property value as $QoI$ for the given distribution parameters.

\subsubsection{Choice of Stochastic Parameters}
\label{intuition_tree_decoder}
All sampled models from $p(\boldsymbol{\omega}; \boldsymbol{\mu}_f,\boldsymbol{\sigma}_{f})$ share the same weights as the pre-trained model for the parameters in $\boldsymbol{\Theta}^D$. The distribution only affects the stochastic parameters in $\boldsymbol{\Theta}^S$. We can construct the active subspace over the entire parameter space, $\boldsymbol{\Theta}$. However, there is no guarantee that the learned subspace would focus on the specific component of the generative model that is closely related to the downstream task. It is intuitive to construct a subspace specifically for those parameters that we aim to modify in the model.

In our work, we have considered the VAE-based generative model for molecular design, particularly JT-VAE \cite{jtvae_paper}, SELFIES-VAE \cite{gao2022sample} and SMILES-VAE \cite{gomez2018automatic}. Given the design points, i.e. set $Q$ in the latent space of the pre-trained VAE model suggested by some generic optimization algorithm $\mathcal{A}$, decoders of the VAE models transform them into molecules. Since we want to get molecules with better properties for the same design points of $Q$ compared to the pre-trained model, we learn the active subspace of the decoders of SELFIES-VAE and SMILES-VAE. In case of JT-VAE, reconstruction of molecules involves two types of decoders. First, the tree decoder predicts a junction tree (JT) from a latent point. Conditioned on this predicted junction tree, the graph decoder constructs the molecular graph by selecting the best arrangement in each node of the junction tree. Out of these two components, the tree decoder plays the pivotal role in deciding the molecular structure as the junction tree contains all coarse information, i.e. which molecular units will be present in the constructed molecule. The graph decoder tracks the fine details of the interconnection between the nodes of the junction tree. Consequently, the tree decoder has broad control over the decision rules for constructing molecules from latent space. Therefore, we construct the active subspace for tree decoder parameters of JT-VAE, as it effectively allows us to control the decision rules for constructing the junction tree from a latent point. 
Our optimization process will try to change those rules, i.e. by changing decoder weights so that a latent point is decoded to a different junction tree leading to a better molecular graph than the one we get using the pre-trained JT-VAE model.
% The encoder counterparts also have some role in the decoding stage. However, on an abstract level, they primarily offer the features or hidden representations upon which decoders base their decisions.
% Intuitively, achieving the desired decision outcome is easier by altering the decision rules rather than the decision variables themselves. 

\subsubsection{Design Space}
\label{design_space}
We are performing the optimization of (\ref{red_opt_problem}) in terms of the active subspace distribution parameters. Since the active subspace parameters drawn from this distribution decide the model instances of the generative model, any large deviation from the posterior distribution may cause the sampled models to behave erroneously on the design points, i.e. leading to invalid chemical structures. So we constrict the design space of our optimization by selecting bounds of $\boldsymbol{\mu}_{f}$ and $\boldsymbol{\sigma}_{f}$ within the neighborhood of the inferred posterior distribution parameters, i.e. $\boldsymbol{\mu}_{\text{post}}$ and $\boldsymbol{\sigma}_{\text{post}}$ as follows:
\begin{align}
\boldsymbol{\mu}_{\text{post}}-3\boldsymbol{\sigma}_{\text{post}} & \leq \boldsymbol{\mu}_{f} \leq \boldsymbol{\mu}_{\text{post}}+3\boldsymbol{\sigma}_{\text{post}} \label{mean_bounds} \\ 
0.75\boldsymbol{\sigma}_{\text{post}} & \leq \boldsymbol{\sigma}_{f} \leq  1.25\boldsymbol{\sigma}_{\text{post}} \label{var_bounds} 
\end{align}

We chose the $3\boldsymbol{\sigma}_{\text{post}}$ half-width around the posterior mean, $\boldsymbol{\mu}_{\text{post}}$, to enable the fine-tuned distribution to navigate within the subspace region aligned with the posterior. The bounds for $\boldsymbol{\sigma}_{f}$ are set to avoid significant variance changes, as this could introduce excess noise in $QoI$ evaluation, potentially hindering the optimization algorithm's performance.

With the above uncertainty guided design space, we put a constraint defined in (\ref{con_eqn}) based on the KL divergence between the fine-tuned and the posterior distribution. By setting the threshold -- $\delta_{KL}$, we can control how far away from the inferred posterior we want to look for a better pool of models. If the design space is already very narrow, then this constraint can be dropped for optimization.
\begin{align}
    KL\left(  p(\boldsymbol{\omega} ; \boldsymbol{\mu}_f,\boldsymbol{\sigma}_{f}) \Vert p(\boldsymbol{\omega} ; \boldsymbol{\mu}_{\text{post}},\boldsymbol{\sigma}_{\text{post}})  \right)  \leq \delta_{KL} \label{con_eqn}
\end{align}

\subsection{Optimization Procedure}
We employ two black-box optimization approaches -- Bayesian optimization and REINFORCE \cite{williams1992simple} to fine-tune the distribution parameters for the optimization problem in \ref{opt_problem}. In the following sections, we discuss how these approaches (details in \cref{appendix: opt_detals}) improve the pre-trained model's $QoI$.

\subsubsection{Bayesian Optimization}
We formulate the optimization problem of (\ref{red_opt_problem}) as a single objective Bayesian optimization (SOBO) task with the KL divergence constraint from (\ref{con_eqn}). To initialize the Gaussian process \cite{williams2006gaussian}-based surrogate model, a small number of candidate solutions, i.e. pairs of  $\left( \boldsymbol{\mu}_{f},\boldsymbol{\sigma}_{f} \right)$ are drawn by applying Sobol' sampler \cite{sobol1967} within the design space defined by (\ref{mean_bounds}) and (\ref{var_bounds}). Then we evaluate these candidates using the $QoI$ function $\phi$ as well as the KL divergence constraint. We train the GP model with the evaluated $QoI$s and constraint slacks for the initial candidates and use it for optimizing the acquisition function (expected improvement in our main experiments) that suggests the next candidate pair in the design space. The $QoI$ and the constraint slack of the suggested pair are similarly evaluated and used to update the GP model. We then repeat the optimization of the acquisition function using the updated surrogate model to get the next candidate to evaluate. This iterative process: optimization of acquisition function and updating GP with new observation -- is repeated until we reach the desired region of $QoI$ or the computational budget for $QoI$ evaluation is exhausted. 

\subsubsection{REINFORCE}
For applying REINFORCE \cite{williams1992simple} in our problem, we consider $p(\boldsymbol{\omega}; \boldsymbol{\mu}_f,\boldsymbol{\sigma}_{f})$ as a policy for the active subspace parameters $\boldsymbol{\omega}$. 
The parameters of the policy network -- $\boldsymbol{\psi }\triangleq (\boldsymbol{\mu}_f,\boldsymbol{\sigma}_{f})$ are first initialized to the active subspace posterior distribution parameters, i.e. $\boldsymbol{\mu}_{\text{post}}$ and $\boldsymbol{\sigma}_{\text{post}}$ respectively. At each iteration of REINFORCE, we draw $M$ samples from current policy network $p(\boldsymbol{\omega}; \boldsymbol{\mu}_f,\boldsymbol{\sigma}_{f})$ and compute the corresponding $QoI$, i.e. $\phi(\boldsymbol{\mu}_f,\boldsymbol{\sigma}_{f},Q)$ following the steps mentioned in \cref{sim_details}. Then we use Adam optimizer \cite{kingma2014adam} with a learning rate $\alpha = 0.005$ to update the policy parameters according to the following update rule:
\begin{align}
    \Delta \boldsymbol{\psi} &= \alpha \phi(\boldsymbol{\mu}_f,\boldsymbol{\sigma}_{f},Q)\pdv{}{ \boldsymbol{\psi}}\left (\sum_{i=1}^{M} \log p(\boldsymbol{\omega}_i; \boldsymbol{\mu}_f,\boldsymbol{\sigma}_{f} ) \right)
\end{align}

\section{Results} \label{Sec:Results}
In this section, we demonstrate the performance of our approach (summarized in \cref{alg:QoI_optimization}) using three VAE models-- JT-VAE, SELFIES-VAE, and SMILES-VAE with two optimization methods-- Bayesian optimization (BO) and REINFORCE. First we elaborate the downstream design tasks in \cref{sim_details} followed by results in \cref{opt_result} showing improvements in molecular properties using our uncertainty-guided fine-tuning over pre-trained models. \cref{as_bias} offers further insights into the active subspaces constructed for each VAE model.

\subsection{Simulation of Downstream Task}
\label{sim_details}
To demonstrate our approach for any set of design points $Q$, we use a random selection strategy as the algorithm $\mathcal{A}$ which draws $1000$ points in the pre-trained VAE's latent space according to $\mathcal{N}(\mathbf{0}, \mathbf{I})$. For simulating $QoI_{\text{PTM}}$, we convert this random collection $Q$ to corresponding molecules using the pre-trained model and predict the property of interest for all unique molecules. $QoI_{\text{PTM}}$ is defined as the average property value of the top $10\%$ samples among those unique molecules. For properties that need to be minimized, the top $10\%$ samples are those with the lowest property values, and the sign of their average is altered to maximize the $QoI$.

For the posterior and fine-tuned distributions over AS parameters, we independently sample $10$ models using the AS parameter distribution and divide the $1000$ latent points of $Q$ equally among them, giving each model $100$ design points to decode. This ensures up to $1000$ unique molecules for a fair comparison with the pre-trained model. We then use the same property predictor on unique molecules that we used for pre-trained model and define the $QoI$ as the average of the top $10\%$ properties for the given distribution parameters. If the pre-trained model or any of the sampled models cannot decode a latent point into a valid molecule, we discard it in the $QoI$ estimation.

\subsubsection{Properties of Interest}
To investigate optimization efficiency over different landscapes of $QoI$s, we consider six target molecular properties: water-octanol partition coefficient (logP), synthetic accessibility score (SAS), natural product-likeness score (NP score) \cite{HARVEY2008894}, inhibition probability against Dopamine receptor D2 (DRD2) \cite{drd2_structure}, c-Jun N-terminal kinase-3 (JNK3) and glycogen synthase kinase-3 beta (GSK3$\beta$) \cite{li2018multi}. We aim to maximize all properties except SAS, where lower values indicate easier synthesizability. Details of predictors for these six properties are provided in \cref{prop_pred}.

\subsection{Optimization over Active Subspace Improves QoI}
\label{opt_result}
For each of the six properties, we applied the Bayesian optimization (BO) and REINFORCE separately to fine-tune the distribution parameters over the active subspace of JT-VAE's tree decoder, SELFIES-VAE decoder, and SMILES-VAE decoder. For BO, we initialized the GP surrogate models by drawing $5$ sample candidates from the design space described in \cref{design_space} and used the expected improvement (EI) acquisition function to optimize the $QoI$ function $\phi$ over 25 BO iterations. For REINFORCE, the policy network was initialized with the posterior distribution parameterized by $\boldsymbol{\mu}_{\text{post}}$ and $\boldsymbol{\sigma}_{\text{post}}$ and updated over $30$ iterations. For both optimization strategies, we limit the number of $QoI$ evaluations to $30$. That means we run each optimization strategy until we reach $30$ $QoI$ evaluations for the given active subspace distribution parameters, ensuring that the $QoI$ evaluation cost remains same for both strategies. In BO, the $QoI$ evaluations using $5$ initial candidates and $25$ subsequent candidates suggested during the $25$ BO iterations constitute a total of $30$ $QoI$ evaluations.
\begin{figure}[!htb]
\vskip -0.05in
\begin{center}
\centerline{\includegraphics[width=0.8\textwidth]{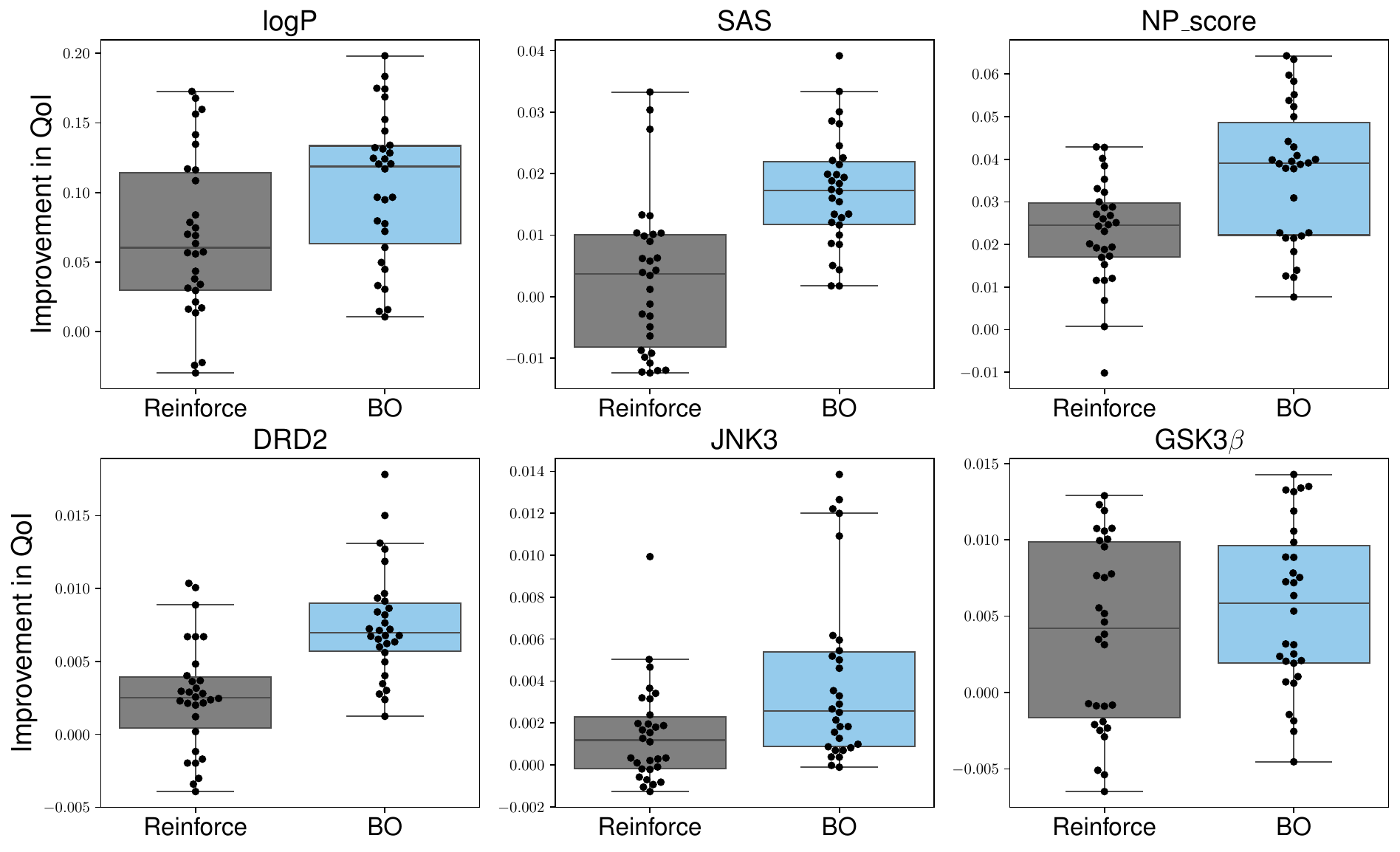}}
\caption{Improvement in $QoI$ relative to the pre-trained JT-VAE model for two optimization methods: BO and REINFORCE. Positive $QoI$ improvement values indicate better $QoI$ than $QoI_{\text{PTM}}$. Each boxplot includes individual $QoI$ improvements for the best fine-tuned distributions found across $10$ different $Q$ sets over 3 optimization trials per optimization method. Some individual observations are horizontally adjusted within each category to remove overlaps among them.}
\label{imprv_QoI_EI}
\end{center}
\vskip -0.2in
\end{figure}
For each property, we run 3 trials of BO and REINFORCE on $10$ independently generated $Q$ sets of design points. Each boxplot in \cref{imprv_QoI_EI} shows the improvement in $QoI$ relative to the pre-trained JT-VAE model over these $30$ runs. The results indicate that the Bayesian optimization approach consistently outperforms the reward-based approach: REINFORCE in achieving larger improvement in $QoI$ for all six properties. \cref{SELFIES_VAE_imprv_QoI_EI,SMILES_VAE_imprv_QoI_EI} in \cref{appendix: selfies_smiles_results} present similar boxplots of $QoI$ improvements for SELFIES-VAE and SMILES-VAE, respectively. For these two models, REINFORCE performs on par with Bayesian optimization in enhancing $QoI$ compared to their respective pre-trained models. For a quantitative comparison, \cref{tab:comparison} reports the $QoI$ values for both the the pre-trained model and the fine-tuned distributions obtained by our approach, showing that our method consistently outperforms the respective pre-trained model in generating molecules with better properties.

\begin{table}[htb]
\caption{Comparison of $QoI$ values obtained by our approach of optimization over the active subspace distribution parameters for the JT-VAE tree decoder, SELFIES-VAE decoder and SMILES-VAE decoder. The $QoI$ for each pre-trained model (PTM) is shown as a baseline to highlight improvements by optimization algorithms: Bayesian optimization (BO) and REINFORCE (R). Each entry is the average $QoI$ over $3$ optimization trials for $10$ different $Q$ sets.}
\label{tab:comparison}
\centering
\scriptsize
\begin{tabular}{@{}clcccccc@{}}
\toprule
Models & \makecell{Pre-trained/ \\ Fine-tuned} &
  \multicolumn{1}{c}{logP ($\uparrow$)} &
  \multicolumn{1}{c}{SAS ($\downarrow$)} &
  \multicolumn{1}{c}{NP score ($\uparrow$)} &
  \multicolumn{1}{c}{DRD2 ($\uparrow$)} &
  \multicolumn{1}{c}{JNK3 ($\uparrow$)} &
  \multicolumn{1}{c}{GSK3$\beta$ ($\uparrow$)} \\ \midrule
\multirow{3}{*}{JT-VAE}      & PTM           & 4.263 (0.084) & 2.034 (0.027) & 0.039 (0.037) & 0.039 (0.008) & 0.061 (0.004) & 0.135 (0.009)       \\
                             & PTM+BO & \textbf{4.367 (0.078)} & \textbf{2.017 (0.029)} & \textbf{0.076 (0.039)} &\textbf{ 0.047 (0.009)} & \textbf{0.065 (0.005)} & \textbf{0.140 (0.008)} \\
                             & PTM+R             & 4.332 (0.093) & 2.031 (0.034) & 0.062 (0.042) & 0.042 (0.010) & 0.062 (0.004) & 0.138 (0.009)        \\ \midrule
\multirow{3}{*}{SELFIES-VAE} & PTM           & 4.741 (0.083) & 2.086 (0.043) & 0.722 (0.035) & 0.039 (0.008) & 0.065 (0.006) & 0.126 (0.007)          \\
                             & PTM+BO & \textbf{5.059 (0.043)} & 2.010 (0.017) & 0.798 (0.028) & 0.064 (0.005) & \textbf{0.076 (0.002)} & 0.146 (0.004)      \\
                             & PTM+R             & 5.044 (0.039) & \textbf{2.001 (0.018)} & \textbf{0.801 (0.021)} & \textbf{0.069 (0.009)} & \textbf{0.076 (0.002)} & \textbf{0.147 (0.004)} \\ \midrule
\multirow{3}{*}{SMILES-VAE}  & PTM           & 4.869 (0.050) & 1.952 (0.024) & 0.194 (0.050) & 0.036 (0.009) & 0.067 (0.005) & 0.117 (0.011)          \\
                             & PTM+BO & 5.050 (0.042) & 1.903 (0.008) & 0.284 (0.022) & 0.059 (0.006) & 0.080 (0.004) & 0.137 (0.006)          \\
                             & PTM+R            & \textbf{5.069 (0.055)} & \textbf{1.900 (0.010)} & \textbf{0.289 (0.027)} & \textbf{0.063 (0.004)} & \textbf{0.081 (0.003)} & \textbf{0.140 (0.005)} \\ \bottomrule
\end{tabular}
\vskip -0.1in
\end{table}

Since Bayesian optimization outperforms REINFORCE in the JT-VAE model, we have conducted additional experiments in JT-VAE case to analyze the effect of a noisy expected improvement acquisition function, the sensitivity of $\delta_{KL}$, effect of active subspace dimension $k$ in Bayesian optimization, and the generalizability of the fine-tuned distribution's impact on the JT-VAE latent space. These results are discussed in \cref{appendix: additional_experiments}.

\subsection{Does Active Subspace have Intrinsic Bias?}
\label{as_bias}
\begin{figure}[htb]
\vskip 0.1in
\begin{center}
\centerline{\includegraphics[width=0.9\columnwidth]{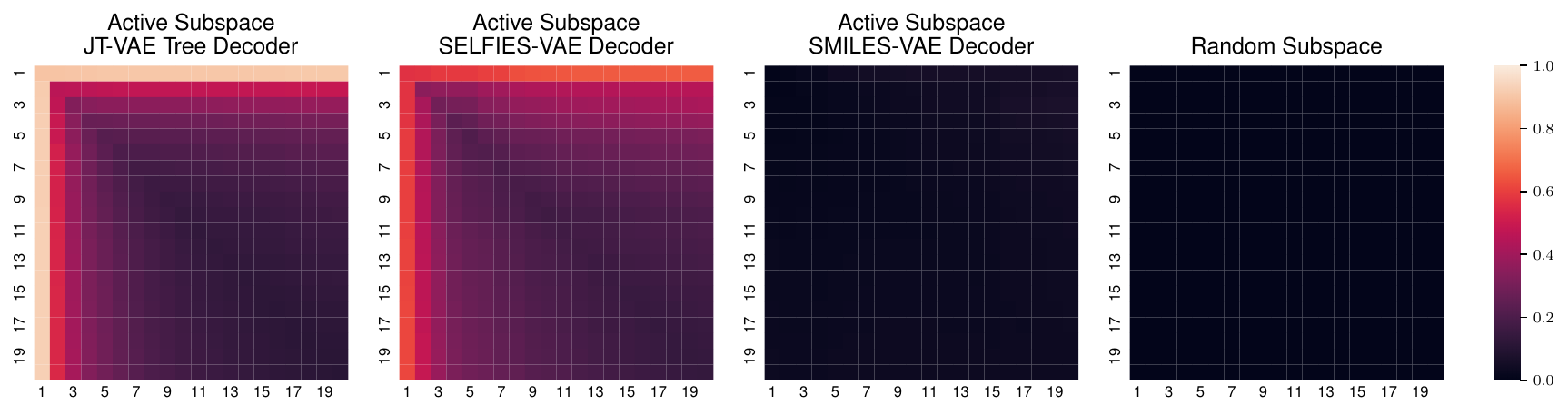}}
\caption{Comparison of subspace similarity between random subspaces and active subspaces for the JT-VAE tree decoder, SELFIES-VAE decoder and SMILES-VAE decoder. Each entry of the normalized subspace similarity is obtained using (\ref{sim_eqn}). In each case, the subspace similarity is calculated between two subspaces generated using two different random seeds.}
\label{AS_vs_Random}
\end{center}
\vskip -0.2in
\end{figure}

We learn the active subspace using a smaller number of gradient samples ($n=100$ in \cref{alg:AS}), with each gradient sample derived from the loss for a single molecule. It's natural to question whether these gradient samples lead to a learned active subspace similar to a random subspace. To investigate this, we compare the subspace similarity between two active subspaces constructed using two random seeds. We use the Grassmann distance-based normalized subspace similarity measure from \cite{hu2021lora}. For two subspaces with projection matrices $\mathbf{P}_1$ and $\mathbf{P}_2$, the subspace similarity is defined as:
\begin{align}
    \operatorname{sim}(\mathbf{P}_1, \mathbf{P}_2, i,j) = \frac{\left\Vert \mathbf{U}^{i\text{T}}_1\mathbf{U}^{j}_2\right\Vert_{F}^{2}}{\min(i,j)} \label{sim_eqn}
\end{align} 
where $\mathbf{U}^{i}_k$ is the first $i$ columns of $\mathbf{P}_k$ after normalization. For our $20$-dimensional active subspaces over the JT-VAE tree decoder, SELFIES-VAE decoder, and SMILES-VAE decoder, \cref{AS_vs_Random} shows this similarity measure for two active subspaces across two random seeds. For reference, we also show the similarity between two random subspaces of the same dimension, where the projection matrices are drawn from a normal distribution. 

Since the random subspaces are independently generated, there is no similarity between them, i.e, the subspace similarity measure is near $0$. While the active subspace is also generated independently across two random seeds, 
the learned projection matrices share a certain degree of intrinsic structure in the cases of JT-VAE tree decoder and SELFIES-VAE decoder, where the first few projection vectors show significant similarity (values closer to $1$). In JT-VAE tree decoder's case, the similarity is particularly pronounced which possibly contributes to the higher improvement in $QoI$ by Bayesian optimization over REINFORCE. This intrinsic bias might be introduced due to the way the JT-VAE tree decoder reconstructs a molecule from the latent space. In contrast, the active subspaces for the SMILES-VAE decoder show negligible similarity, despite having a similar architecture to SELFIES-VAE, except for different molecular representations. This difference suggests that the SMILES-VAE's pre-trained weights may reside in a very sharp loss surface. Perturbing the weights (for active subspace construction) in any direction might lead to a large change in the loss value, resulting in a subspace similar to a random subspace. These observations indicate the importance of robust input representations, as seen in JT-VAE and SELFIES-VAE models, for learning an effectively low-dimensional active subspace. Furthermore, it may be possible to construct the subspace that favors optimum regions of $QoI$, offering a promising direction for future work.

\section{Related Work}
\label{Sec:Related_work}
% \hl{rephrased. check:} 
\cite{mackay1991bayesian} introduced the concept of data-dependent effective dimensionality of neural network parameter space in Bayesian framework. The experiments of \cite{maddox2020rethinking} demonstrated the existence of many directions within the neighborhood of trained neural network weights where predictions remain unchanged. Several works \cite{frankle2018LOT,blundell2015weight, Louizos-et-al-2017, Jantre-et-al-2023, jantre2023comprehensive} utilized this concept to to compress over-parameterized neural networks by pruning. Furthermore, this low dimensionality in parameter space enables scalable uncertainty quantification through various subspace inference techniques \cite{izmailov2020subspace, jantre2024active}. In our work, we chose the active subspace approach \cite{jantre2024active} over other methods since it allows learning the subspace without retraining or modifying the architecture of the pre-trained model.

Previous efforts to fine-tune GMD models have been limited to using small, select molecule sets that fit specific design criteria. For example, \cite{blaschke2021finetune} fine-tuned the REINVENT model \cite{blaschke2020reinvent} to improve its ability to recognize molecules with multi-target attributes via transfer learning, i.e. retraining the pre-trained model with a pool of multi-target molecules. In contrast, our approach aims to adapt the generative model based on downstream task performance. This problem is conceptually similar to the work by \cite{krupnik2023fine}, where they update the pre-trained generative model parameters to generate samples aligned with the observed data from a robotics task simulator.

\section{Conclusion and Discussion}
\label{Sec:Conclusion}
We introduced an uncertainty-guided fine-tuning approach that leverages a pre-trained VAE-based generative model's low-dimensional active subspace to quantify model uncertainty which we further explore enhancing the performance in downstream molecular design tasks. Our method showed significant improvements over pre-trained models for six molecular properties for three VAE variants: JT-VAE, SELFIES-VAE, and SMILES-VAE. By leveraging black-box optimization, our approach can fine-tune the generative model to improve predicted properties using any property predictor, whether ML or mechanistic. Additionally, our Bayesian optimization framework can naturally extend to multi-objective optimization when multiple molecular properties are of interest. Our results highlight the varying impacts of the models derived from the active subspace distribution on different molecular properties, suggesting a need for objective-guided active subspace development. Moreover, active subspace inference enables formal uncertainty quantification of generative models and property predictors, which may offer insights into data-driven generative molecular discovery.
% allows one to utilize any property predictor, either ML or mechanistic model, to fine-tune the generative model to enhance the corresponding property of the molecules it generates. 
% Furthermore, our Bayesian optimization framework naturally paves the way for multi-objective optimization of the generative model when multiple molecular properties are of interest.
% Our experiments reveal that the models derived from the active subspace posterior distribution affect the $QoI$ of design points differently for each property. This motivates future work on objective-guided active subspace development. Furthermore, the success of fine-tuning depends on the pre-trained GMD model quality. If the pre-trained model does not capture underlying molecular generation rules, improvements in design points via an active subspace approach is unlikely. To this end, a future research direction could involve integrating our proposed approach with iterative refinement methods~\cite{tripp2020sample, yang2020improving, abeer2022multi} to improve the generative model's sampling efficiency. 
% Our experiments reveal that the models derived from the active subspace distribution affect the $QoI$ of design points differently for each property. 
% This motivates future work on objective-guided active subspace development. Moreover, active subspace inference enables formal uncertainty quantification of generative models for molecular design and neural network-based property predictors, which may provide further insights into data-driven generative molecular discovery.

\textbf{Limitations.} The success of our proposed fine-tuning approach hinges on the quality of pre-trained generative model, as we learn the active subspace posterior distribution parameters over which the design space is defined, by perturbing the pre-trained weights. If the pre-trained model fails to capture underlying molecular generation rules, our active subspace-based method  is unlikely to yield improvements in design. One potential future direction is to integrate our approach with iterative refinement methods~\cite{tripp2020sample, yang2020improving, abeer2022multi} to enhance the generative model's sampling efficiency. Moreover, our fine-tuning method can be misguided \cite{martinelli2023leveraging} if the surrogate models used for $QoI$ feedback do not accurately represent the ground truth feedback. To mitigate this, one can leverage our black-box treatment to adopt a risk-aware optimization approach for robust fine-tuning.

\textbf{Broader Impacts.} 
As generative models gain popularity, biopharmaceutical companies are increasingly turning to generative design for applications like small molecule drugs and protein sequence design. Our approach offers a modular extension to their current pipeline, allowing them to adapt generative molecular design models for specific developability requirements in different downstream tasks. This speeds up the fine-tuning of these models for new design criteria, as we update model parameters based on the feedback from new tasks rather than waiting for a large number of samples that meet the requirements. We do not foresee any major negative impacts of our work, but policymakers setting design requirements of the downstream task should consider their direct implications. Additionally, our fine-tuning approach has the risk of misuse in potentially expediting the identification of ingredients necessary for designing more destructive chemical weapons.

\section{Acknowledgement}
This research was supported by funding from the Advanced
Scientific Computing Research program in the United States Department of Energy’s Office of Science under grant 0000269227  and  
projects B\&R\# KJ0402010 and FWP\# CC125.

\bibliography{references}
\bibliographystyle{unsrt_abbrv_custom}
\newpage
\section{Appendix}

% Optionally include supplemental material (complete proofs, additional experiments and plots) in appendix.
% All such materials \textbf{SHOULD be included in the main submission.}

\subsection{Generative Model for Molecular Design}
\label{appendix: GMD}
\subsubsection{JT-VAE}
We consider the Junction Tree Variational Autoencoder (JT-VAE) from \cite{jtvae_paper} as one of the generative models for molecular design. The encoder which consists of two components, i.e. tree and graph encoders converts the input molecule to $56$ dimensional latent space. Specifically, the tree encoder projects the junction tree of the molecule to the corresponding $28$ dimensional junction tree embedding and the graph encoder finds the $28$ dimensional molecular graph embedding using the molecular graph. During reconstruction, the tree decoder constructs the junction tree from the junction tree embedding. Then the graph decoder reconstructs the molecular graph conditioned on the predicted junction tree from the tree decoder.

We used the pre-trained JT-VAE along with its training and validation dataset from \url{https://github.com/cambridge-mlg/weighted-retraining} \cite{tripp2020sample}.

\subsubsection{SELFIES-VAE and SMILES-VAE}
Both SELFIES-VAE and SMILES-VAE have similar neural architecture except for the input representation for molecules: Simplified Molecular-Input Line-Entry System (SMILES) \cite{weininger1988smiles} and SELFreferencIng
Embedded Strings (SELFIES) \cite{krenn2020self} respectively. 
In both the models, encoder embeds an input molecule into a $128$ dimensional latent space from which corresponding decoder reconstructs a molecule in an autoregressive fashion.
The pre-trained models and corresponding training datasets used in our work are taken from \url{https://github.com/wenhao-gao/mol_opt} \cite{gao2022sample}.

\subsection{Property Predictors}
\label{prop_pred}
The RDKit package \cite{landrum2013rdkit} is used for predicting logP, SAS, and NP score \cite{NP_score_paper} from the SMILES of the molecules.  The oracles for JNK3 and GSK3$\beta$ provided in Therapeutics Data Commons \cite{Huang2021tdc} predict the corresponding inhibition probability using random forest classifiers.  Finally, as the predictor of the DRD2 inhibition probability, we trained the support vector machine (SVM) classifier proposed in \cite{olivecrona2017} using their dataset \url{https://github.com/MarcusOlivecrona/REINVENT/releases/download/v1.0.1/data.tar.gz}.

\subsection{Active  Subspace}
\label{appendix: AS}
\subsubsection{Learning Active Subspace}\label{learning_AS}
\begin{algorithm}[htb]
\caption{Active subspace inference for VAE based generative model}
\label{alg:AS}
\begin{algorithmic}[1]
    \STATE {\bfseries Input:} loss function $\mathcal{L}$ used to train $\mathcal{M}_{\boldsymbol{\theta}_0}$, pre-trained model weights $\btheta_0 = \left[ \btheta^S_0, \btheta^D_0 \right]$, training dataset $\mathcal{D}$ of pre-trained model, number of gradient samples $n$, active subspace dimension $k$, perturbation standard deviation $\sigma_0$.
    \FOR {$j= 1,2,\dots, n$}
    \STATE Sample an input molecule $\boldx_j \in \mathcal{D}$
    \STATE Sample $\btheta^S_j \sim \calN(\btheta^S_0,\sigma_0^2\mathbf{I})$ 
    \STATE Compute gradients: $\nabla_{\btheta^S_j} \mathcal{L}(\mathcal{M}_{\boldsymbol{\theta}_j},\boldx_j)$ where $\btheta_j = \left[ \btheta^S_j, \btheta^D_0 \right]$
    \ENDFOR
    \STATE Uncentered covariance matrix of loss gradients approximated by MC sampling: \\ \centerline{$\hat{\calC} = \frac{1}{n} \sum_{j=1}^n (\nabla_{\btheta^S_j} \mathcal{L}(\mathcal{M}_{\boldsymbol{\theta}_j}, \boldx_j)) (\nabla_{\btheta^S_j} \mathcal{L}(\mathcal{M}_{\boldsymbol{\theta}_j},\boldx_j))^T$}
    \STATE Eigendecomposition of $\hat{\calC}$
    \STATE \textbf{Active subspace} -- spanned by the eigenvectors corresponding to $k$ largest eigenvalues of $\hat{\calC}$
    \STATE Approximate posterior distribution over subspace parameters $\boldsymbol{\omega}$ through VI.
    \STATE Draw $M$ samples of active subspace parameters: $\boldsymbol{\omega}_m \sim p(\boldsymbol{\omega}|\calD) \quad \text{where,} \enskip m \in \{1,\cdots, M\}$  
    \STATE Compute VAE model weights for each sample: $\btheta_m = \left[\btheta^S_0 + \mathbf{P} \boldsymbol{\omega}_m, \btheta^D_0 \right]$
\end{algorithmic}
\end{algorithm} 

\begin{algorithm}[htb]
\caption{Uncertainty-guided fine-tuning of VAE based generative model for improving $QoI$}
\label{alg:QoI_optimization}
\begin{algorithmic}[1]
    \STATE {\bfseries Input:} VAE model $\mathcal{M}_{\boldsymbol{\theta}_0}$, pre-trained model weights $\btheta_0 = \left[ \btheta^S_0, \btheta^D_0 \right]$, set of candidate latent points $Q$ found by some arbitrary algorithm $\mathcal{A}$.
    \STATE Construct active subspace $\boldsymbol{\Omega}$ for stochastic parameters $\btheta^S \in \boldsymbol{\Theta}^S$  (\cref{alg:AS}).
    \STATE Approximate posterior distribution for $\boldsymbol{\omega} \in \boldsymbol{\Omega}$ with  $p(\boldsymbol{\omega}; \boldsymbol{\mu}_\text{post}, \boldsymbol{\sigma}_\text{post})$ via variational inference.
    \STATE Define the design space guided by model uncertainty parameters: $\boldsymbol{\mu}_\text{post}, \boldsymbol{\sigma}_\text{post}$. (\cref{design_space})
    \STATE Run Bayesian optimization/ REINFORCE to solve optimization problem in (\ref{red_opt_problem}) for improving $QoI$ of the latent points in given $Q$ set. 
\end{algorithmic}
\end{algorithm} 

We randomly sampled $100$ molecules from the training dataset of each VAE models considered in our work, and followed \cref{alg:AS} to compute the gradient of the training loss function for each sample (of perturbed model parameters and one input molecule) for constructing the active subspace over the stochastic set of parameters which contains the parameters of tree decoder in JT-VAE, and decoder in cases of SELFIES-VAE and SMILES-VAE. Note during the forward pass of each sample, we did not change any parameters outside of stochastic parameters. We constructed a $20$ dimensional active subspace with perturbation standard deviation $\sigma_0=0.1$ for JT-VAE tree decoder ($2756131$ parameters) and $\sigma_0=0.01$ for SELFIES-VAE decoder ($4419177$ parameters) and SMILES-VAE decoder ($4281894$ parameters). As shown in \cref{AS_vs_Random}, the JT-VAE tree decoder's subspace even may have a lower rank. However, we found that the lowest singular value out of $20$ dimensions was not very small. Hence, we went ahead with $20$ dimensional active subspace in our main experiments. In \cref{appendix: as_dim_5} we have included additional results with $5$ dimensional active subspace.

We used the training dataset of the VAE model to perform variational inference to approximate the posterior distribution over active subspace parameters. We applied the Adam optimizer \cite{kingma2014adam} with a learning rate of $0.001$  to find the approximate mean and standard deviation over $20$ dimensional active subspace parameters by minimizing the combined loss of VAE training loss (that includes the reconstruction loss and KL divergence of VAE) with KL divergence loss between approximated posterior distribution and prior distribution over AS. For the prior distribution over AS parameters, a multivariate normal distribution with zero mean and $5$ standard deviation is used.

\subsubsection{Optimization over Distribution of Active Subspace Parameters}
\label{appendix: opt_detals}
Gaussian process models used in Bayesian optimization had Matern kernel with smoothness parameter, $\nu = 2.5$. Each design variable of BO was rescaled to be within $[0,1]$ and the objective or outcome for each sample design was normalized to have zero mean and unit variance. The Bayesian optimization pipeline is implemented in BoTorch \cite{balandat2020botorch}.

At each update stage of REINFORCE, we additionally clamp the policy network parameters to the corresponding lower and upper bounds defined in \cref{mean_bounds,var_bounds}. To enforce the KL divergence based constraint in REINFORCE, we replace the evaluated $\phi(\boldsymbol{\mu}_f,\boldsymbol{\sigma}_{f},Q)$ by a large negative number whenever $\boldsymbol{\mu}_f,\boldsymbol{\sigma}_{f}$ does not satisfy the constraint defined in (\ref{con_eqn}).

For our experiments in the main text, we empirically set the KL-divergence threshold, $\delta_{KL}$ of (\ref{con_eqn}) to be around $70\%$ of the largest KL-divergence value observed at the bounds of design space defined by (\ref{mean_bounds}) and (\ref{var_bounds}). 
We have included experimental results with different values of $\delta_{KL}$ in \cref{appendix: delta_kl}.
Depending on the bounds on $\boldsymbol{\sigma}_f$, the KL-divergence at the boundary of the design space can be very large. In that case, this threshold needs to be adjusted due to the trade-off between finding a better solution within the limited number of BO iterations without moving far away from the posterior active subspace.

\clearpage

\subsection{Improvement Trend in SELFIES-VAE and SMILES-VAE}
\label{appendix: selfies_smiles_results}
\begin{figure}[!htb]
\vskip 0.2in
\begin{center}
\centerline{\includegraphics[width=0.82\textwidth]{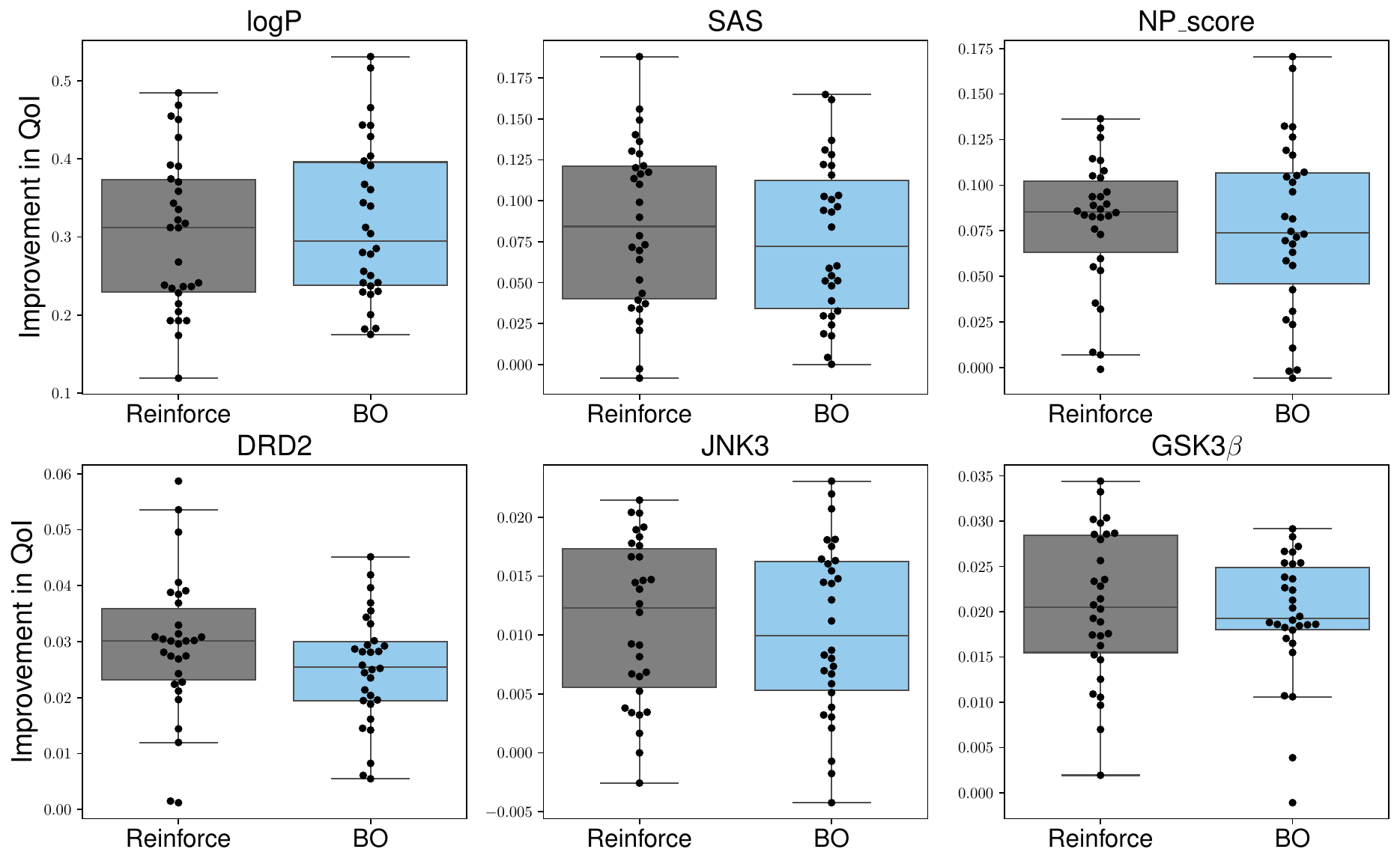}}
\caption{Improvement in $QoI$ relative to the pre-trained SELFIES-VAE model for two optimization methods: BO and REINFORCE. Positive $QoI$ improvement values indicate better $QoI$ than $QoI_{\text{PTM}}$. 
Each boxplot includes individual $QoI$ improvements for the best fine-tuned distributions found across $10$ different $Q$ sets over 3 optimization trials per optimization method. Some individual observations are horizontally adjusted within each category to remove overlaps among them.}
\label{SELFIES_VAE_imprv_QoI_EI}
\end{center}
\vskip -0.2in
\end{figure}

\begin{figure}[!htb]
\vskip 0.2in
\begin{center}
\centerline{\includegraphics[width=0.82\textwidth]{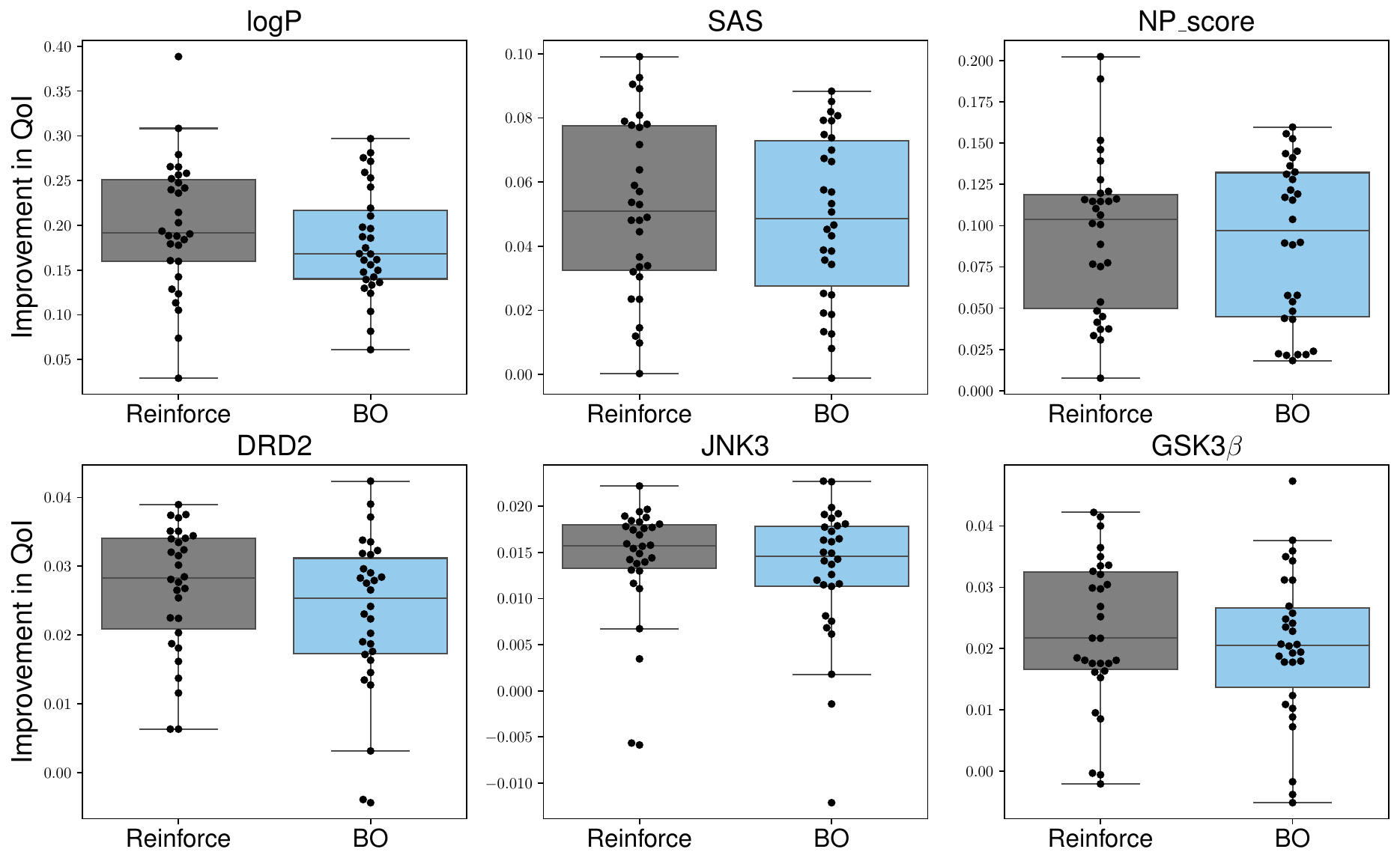}}
\caption{Improvement in $QoI$ relative to the pre-trained SMILES-VAE model for two optimization methods: BO and REINFORCE. Positive $QoI$ improvement values indicate better $QoI$ than $QoI_{\text{PTM}}$. 
Each boxplot includes individual $QoI$ improvements for the best fine-tuned distributions found across $10$ different $Q$ sets over 3 optimization trials per optimization method. Some individual observations are horizontally adjusted within each category to remove overlaps among them.}
\label{SMILES_VAE_imprv_QoI_EI}
\end{center}
\vskip -0.2in
\end{figure}

\subsection{Additional Experiments for JT-VAE}
\label{appendix: additional_experiments}

\subsubsection{Relative Improvement Trend for Different Acquisition Functions}
\label{appendix_rel_imprv}
\begin{figure*}[ht]
\vskip 0.2in
\begin{center}
\centerline{\includegraphics[width=\columnwidth]{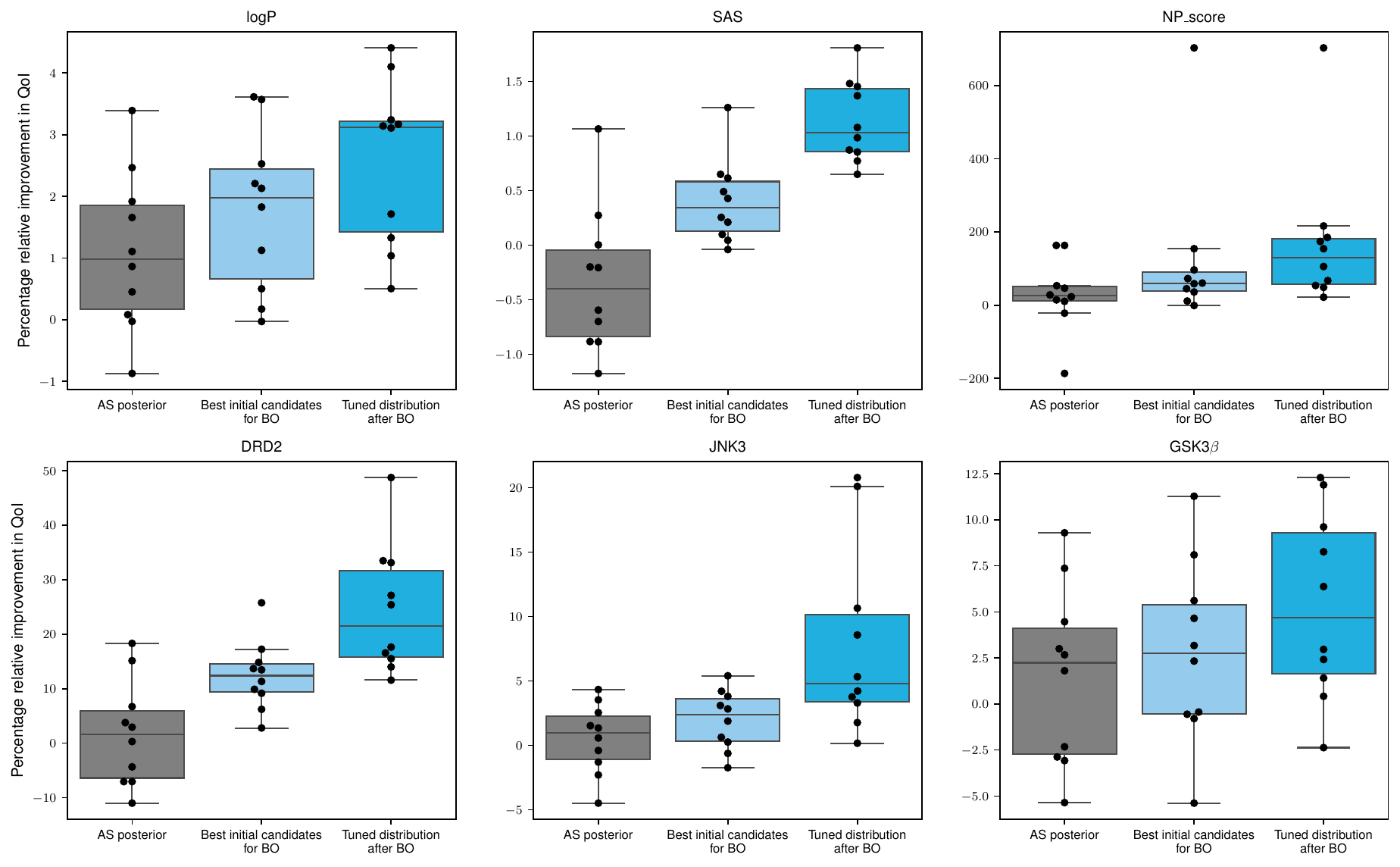}}
\caption{Percentage relative improvement in $QoI$ relative to pre-trained JT-VAE at different stages of optimization over distribution of active subspace parameters using expected improvement (EI) as the acquisition function. For each $QoI$, the boxplots (from left to right) are showing the relative improvement due to posterior of active subspace parameters, best candidate distributions out of random samples needed for initiating the BO, and best solution found after BO iterations respectively. Some observations are horizontally adjusted within each category to remove overlaps among them.}
\label{rel_imprv_QoI_EI}
\end{center}
\vskip -0.2in
\end{figure*}

\begin{figure*}[ht]
\vskip 0.2in
\begin{center}
\centerline{\includegraphics[width=\columnwidth]{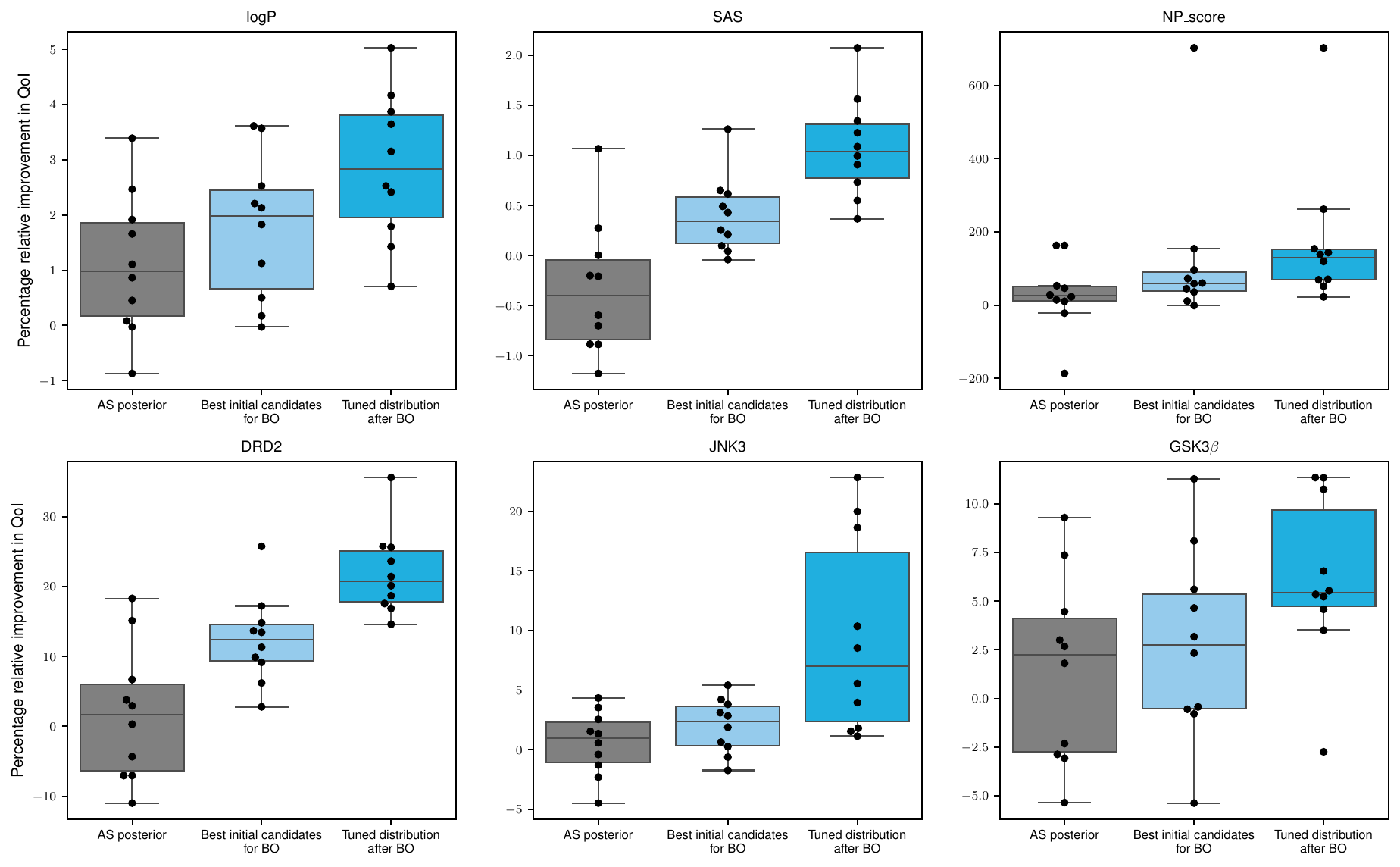}}
\caption{Percentage relative improvement in $QoI$ relative to pre-trained JT-VAE at different stages of optimization using noisy expected improvement (NEI) as the acquisition function. For each $QoI$, the boxplots (from left to right) are showing the relative improvement due to posterior of active subspace parameters, best candidate distributions out of random samples needed for initiating the BO, and best solution found after BO iterations respectively. Some observations are horizontally adjusted within each category to remove overlaps among them.}
\label{rel_imprv_QoI_NEI}
\end{center}
\vskip -0.2in
\end{figure*}

Each subplot in \cref{rel_imprv_QoI_EI,rel_imprv_QoI_NEI} (corresponding to EI and NEI acquisition function respectively) shows three boxplots of relative $QoI$ improvement with respect to the pre-trained JT-VAE model for the corresponding property of interest. The first one is the improvement due to the posterior distribution, i.e. $QoI_{\text{post}} -QoI_{\text{PTM}}$. The other two contain the improvement due to the best fine-tuned distribution, $p(\boldsymbol{\omega} ; \boldsymbol{\mu}_f,\boldsymbol{\sigma}_{f})$ at initialization and final iteration steps of Bayesian optimization. The initial best $QoI$ is selected from the $5$ initialization candidates drawn randomly from the design space. The best pair of $\left(\boldsymbol{\mu}_f,\boldsymbol{\sigma}_{f}\right)$ is tracked over the BO iterations and the third boxplot shows its corresponding improvement with respect to the $QoI_{\text{PTM}}$. We compute the relative improvement by dividing the improvement from $QoI_\text{PTM}$ by $\lvert QoI_\text{PTM} \rvert$. In this additional experiment, we have used $50$ BO iterations for both acquisition functions.

Across all six properties, Bayesian optimization with EI was able to improve the $QoI$s for the given set $Q$. Even in the case of SAS and DRD2 inhibition probability where the posterior distribution tends to give $QoI_\text{post}$ lower than $QoI_\text{PTM}$, we managed to reach suitable fine-tuned distribution with better $QoI$. For GSK3$\beta$, in nine out of ten $Q$s, the fine-tuned distribution outperformed the pre-trained model. Moreover, the small difference in improvement between the initial and final candidates (after BO) for logP and NP score suggests that the active subspace favors the optimization direction for these properties. This is expected since the posterior distribution shows a positive impact on these two $QoI$s before applying the Bayesian optimization. Even for properties like JNK3, where the posterior shows no significant improvement in $QoI$ on average, Bayesian optimization successfully yielded a better pool of models.

We also observed a similar trend in optimization trials for the same $10$ sets using the noisy expected improvement (NEI) acquisition function  (\cref{rel_imprv_QoI_NEI}). Specifically for GSK3$\beta$, using NEI resulted in a more favorable improvement trend.

\clearpage

\subsubsection{Sensitivity of $\delta_{KL}$}
\label{appendix: delta_kl}
The threshold -- $\delta_{KL}$ in \cref{con_eqn} defines the valid design space within the neighborhood of the posterior distribution over the active subspace parameters. We have considered three values: ${27,45,63}$ for $\delta_{KL}$ where $63$ is approximately $70\%$ of the largest KL-divergence value observed at the bounds of design space defined by (\ref{mean_bounds}) and (\ref{var_bounds}). We run Bayesian optimization with expected improvement acquisition function for $50$ iterations for optimization of $QoI$ for $10$ $Q$ sets. 

\begin{figure*}[ht]
\vskip 0.2in
\begin{center}
\centerline{\includegraphics[width=\columnwidth]{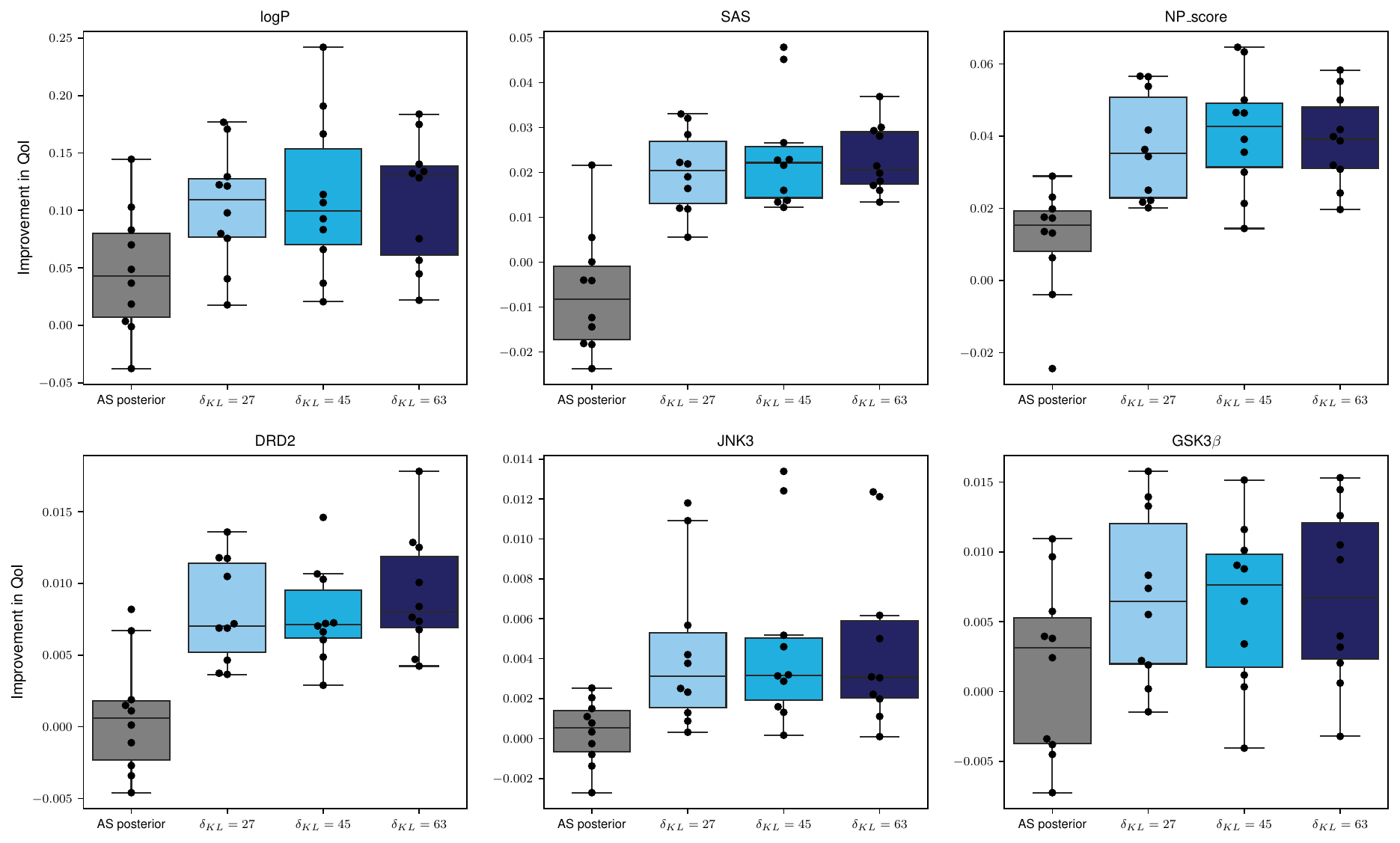}}
\caption{Improvement in $QoI$ relative to pre-trained JT-VAE model by $20$ dimensional active subspace posterior (first boxplot) and fine-tuned distribution found using expected improvement (EI) as the acquisition function for three different $\delta_{KL}$ values.
Positive $QoI$ improvement values indicate better $QoI$ than $QoI_{\text{PTM}}$. 
Each boxplot includes individual $QoI$ improvements for the best fine-tuned distributions found across $10$ different $Q$ sets. Some individual observations are horizontally adjusted within each category to remove overlaps among them.}
\label{diff_thr_imprv_QoI_EI}
\end{center}
\vskip -0.2in
\end{figure*}

In \cref{diff_thr_imprv_QoI_EI} we show the improvement in $QoI$ relative to pre-trained JT-VAE model by $20$ dimensional active subspace posterior (first boxplot) and fine-tuned distribution found using Bayesian optimization for three different values of $\delta_{KL}$. For different properties, we can see that active subspace distribution affect the $QoI$ of design points differently. For example, models sampled from active subspace posterior distribution are performing worse in SAS than pre-trained JT-VAE. And larger $\delta_{KL}$ provides better improvement trend for this property since large value of $\delta_{KL}$ allows exploration farther away from the posterior distribution parameters.

\clearpage

\subsubsection{Leveraging Lower Rank of Active Subspace in JT-VAE}
\label{appendix: as_dim_5}
In \cref{as_bias}, we showed that the $20$ dimensional active subspace of JT-VAE tree decoder is intrinsically low-dimensional. Here, we investigate whether we can get similar trend for improvement in $QoI$ using lower number of  active subspace dimensions instead of $20$. Specifically, we have constructed $5$ dimensional active subspace for JT-VAE tree decoder, and approximated the corresponding posterior distribution via variational inference. Then we perform Bayesian optimization to find fine-tuned distribution over this $5$ dimensional active subspace for improving the $QoI$ for all six properties for same $10$ $Q$ sets as we have done in our experiment with $20$ dimensional active subspace. 

\begin{figure*}[ht]
\vskip 0.2in
\begin{center}
\centerline{\includegraphics[width=\columnwidth]{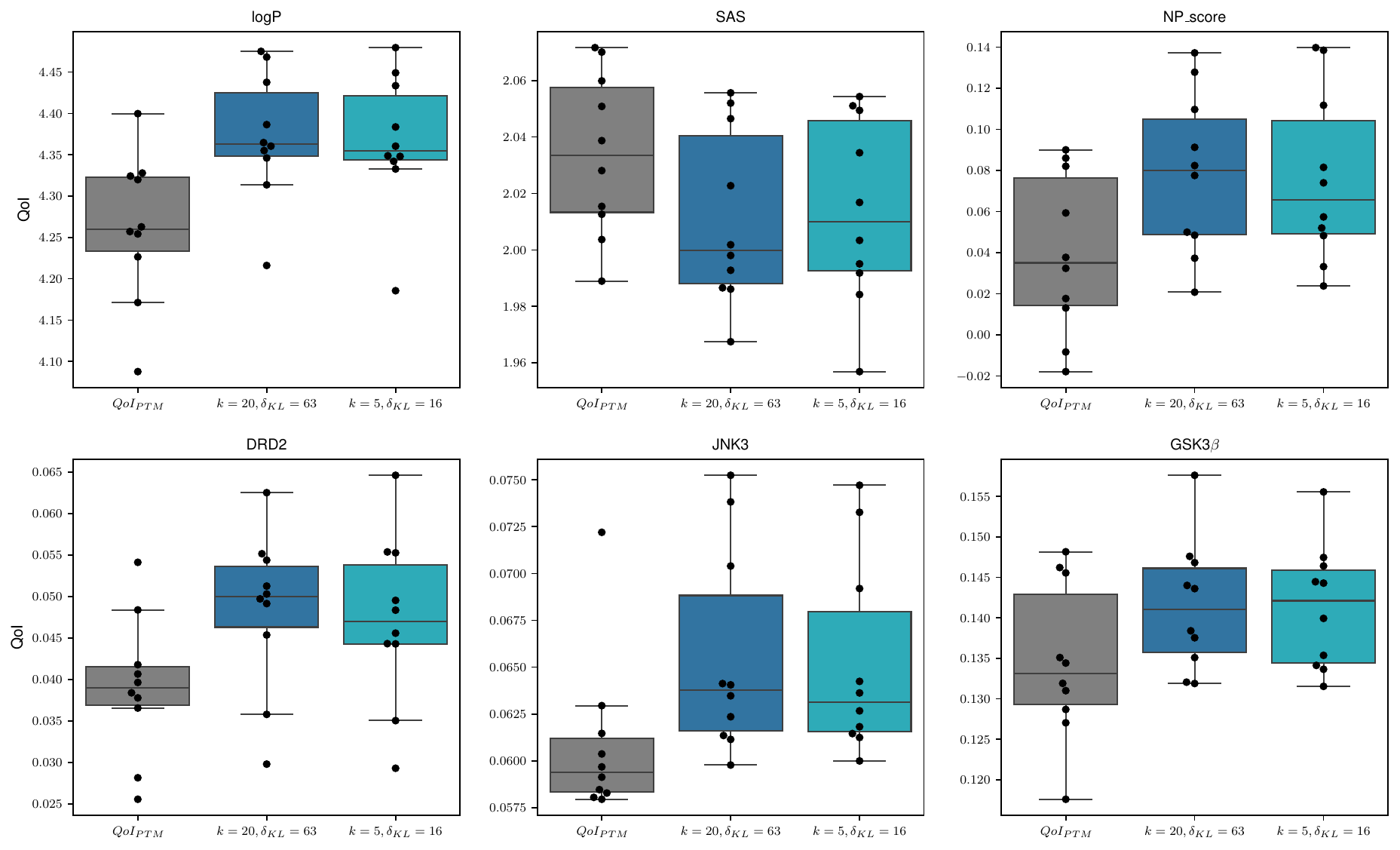}}
\caption{$QoI$ values for pre-trained JT-VAE and fine-tuned distribution for $20$ and $5$ dimensional active subspace parameters found using Bayesian optimization with expected improvement acquisition (EI) function. 
The different values for $\delta_{KL}$ correspond to $70\%$ of the largest KL-divergence value observed at the bounds of design space for $20$ and $5$ dimensional active subspaces.
For SAS, we show the top $10\%$ average property values without the sign alteration mentioned in \cref{sim_details}.
Some observations are horizontally adjusted within each category to remove overlaps among them.}
\label{diff_as_dim_QoI_EI}
\end{center}
\vskip -0.2in
\end{figure*}

In \cref{diff_as_dim_QoI_EI}, we present the $QoI$ values corresponding to fine-tuned distribution over $5$ dimensional active subspace. For reference, we also show the $QoI$ values for the pre-trained JT-VAE model and $20$ dimensional fine-tuned distribution. The trend of improved $QoI$ (maximization for all properties except SAS) is indeed similar between $5$ and $20$ dimensional cases which supports the evidence of lower rank of active subspace in \cref{as_bias}. Note, for both active subspace cases, Bayesian optimization with expected improvement acquisition function is performed for $50$ iterations. For both cases, values of $\delta_{KL}$ are around the $70\%$ of the largest KL-divergence value observed at the bounds of design space for $20$ and $5$ dimensional active subspaces.

\clearpage

\subsubsection{Does Bayesian Optimization Capture General Direction of Improvement in JT-VAE? }
\label{other_Qs}
In \cref{opt_result}, we showed that Bayesian optimization identifies pairs of $\left( \boldsymbol{\mu}_{f},\boldsymbol{\sigma}_{f} \right)$ corresponding to fine-tuned distribution over the active subspace parameters for each of the $10$ $Q$ sets, that enhance $QoI$ compared to the pre-trained model. We examine if those $10$ pairs of $\left( \boldsymbol{\mu}_{f},\boldsymbol{\sigma}_{f} \right)$ corresponding to $10$ $Q$ sets can enhance the $QoI$s for any arbitrary set of design points, $Q'$, for which we did not perform the optimization over active subspace. 

\begin{figure*}[htb]
\vskip 0.2in
\begin{center}
\centerline{\includegraphics[width=\columnwidth]{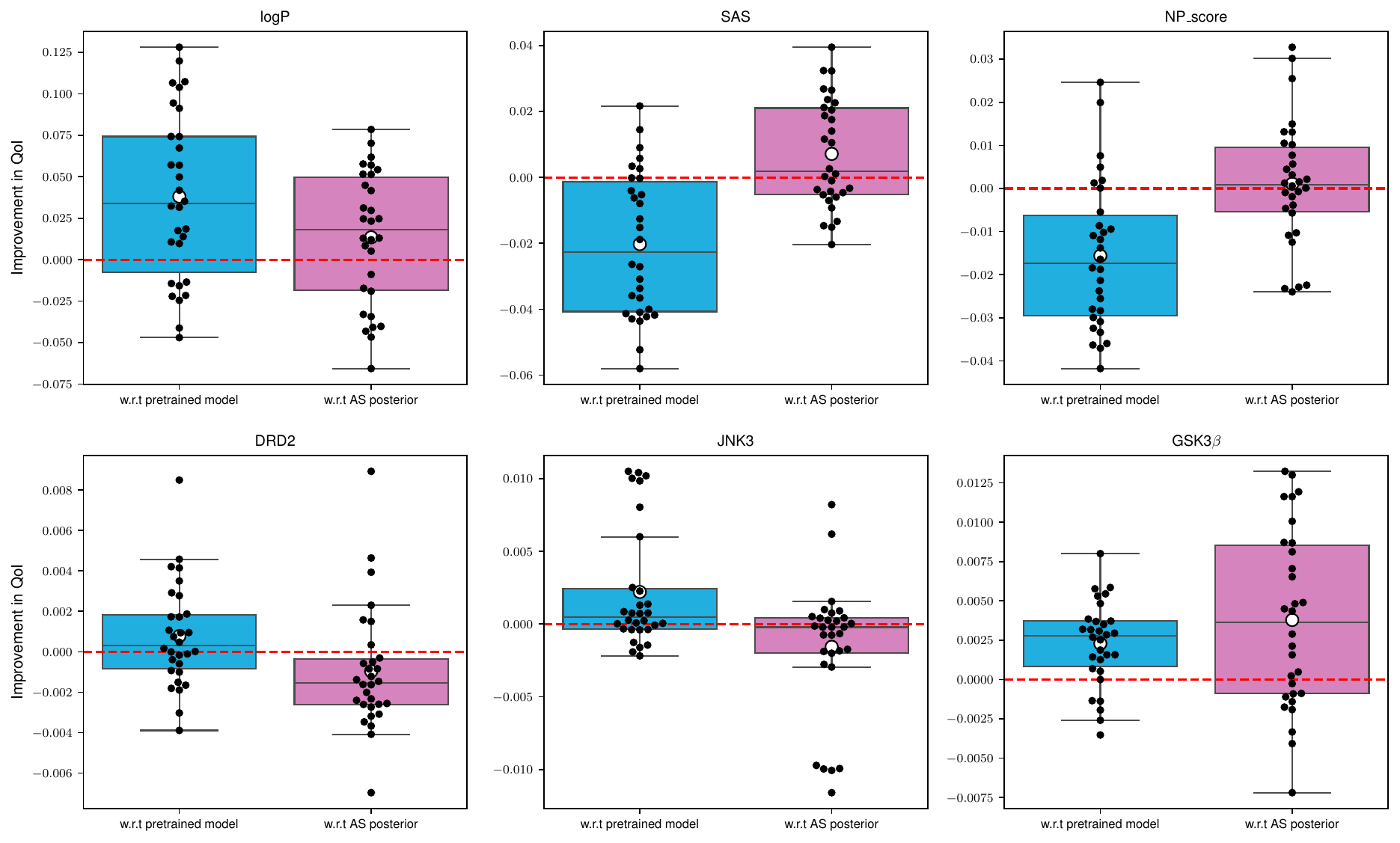}}
\caption{Improvement in $QoI$ for 3 design sets $Q'$ due to $10$ fine-tuned distributions found using expected improvement (EI). $Q'$s are different from the $Q$ sets used in $10$ BO trials. In each subplot, the boxplots (from left to right) are for the improvement with respect to pre-trained JT-VAE model and active subspace posterior distribution respectively. The white circle indicates the average over all $30$ observations. Some observations are horizontally adjusted within each category to remove overlaps among them.}
\label{imprv_QoI_EI_otherQs}
\end{center}
\vskip -0.2in
\end{figure*}

We generate three $Q'$ sets independently by setting different random seeds than the ones used for the $10$ $Q$ sets used in optimization. Next for each $Q'$, we compute the $QoI_\text{PTM}$ and $QoI_\text{post}$. \cref{imprv_QoI_EI_otherQs} has two boxplots for each property showing the improvement trend in each $Q'$ from $QoI_\text{PTM}$ and $QoI_\text{post}$ respectively by using $10$ pairs of $\left( \boldsymbol{\mu}_{f},\boldsymbol{\sigma}_{f} \right)$ found by the BO that uses EI as the acquisition function. 

The first boxplot (left) of each subplot in \cref{imprv_QoI_EI_otherQs} shows the improvement relative to pre-trained JT-VAE performance on three independently generated sets $Q'$ if we use each of these $10$ fine-tuned distributions to generate molecules. In the other boxplot (right) of each subplots in \cref{imprv_QoI_EI_otherQs}, we show the improvement due to these fine-tuned distributions over the active subspace posterior distribution. Note, the fine-tuned distributions used in \cref{imprv_QoI_EI_otherQs} are found by using EI over $50$ BO iterations.

\begin{figure*}[htb]
\vskip 0.2in
\begin{center}
\centerline{\includegraphics[width=\columnwidth]{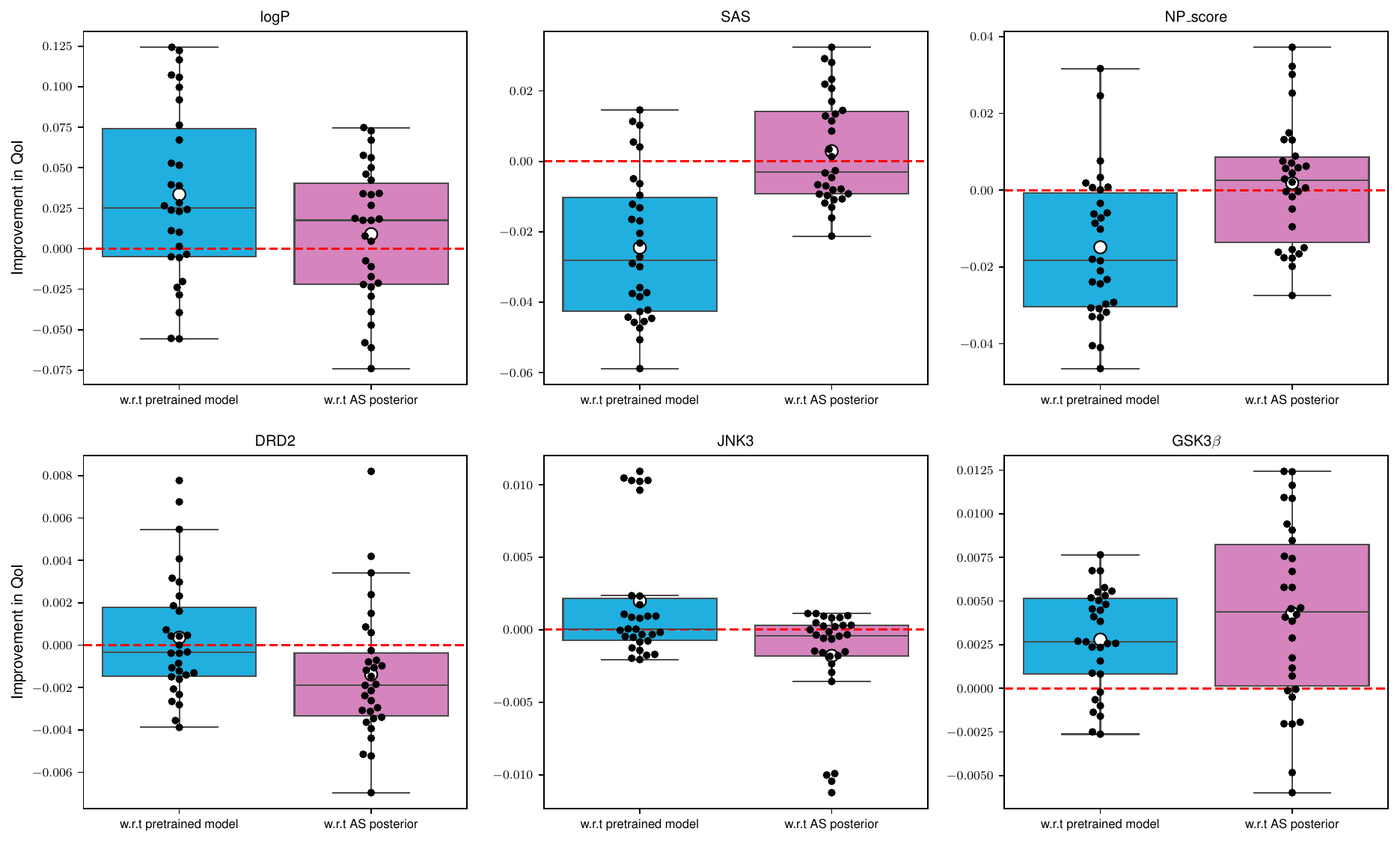}}
\caption{Improvement in $QoI$ for 3 design sets $Q'$ due to $10$ fine-tuned distributions found using noisy expected improvement (NEI). $Q'$s are different from the $Q$ sets used in $10$ BO trials. In each subplot, the boxplots (from left to right) are for the improvement with respect to pre-trained JT-VAE model and active subspace posterior distribution respectively. The white circle indicates the average over all $30$ observations. Some observations are horizontally adjusted within each category to remove overlaps among them.}
\label{imprv_QoI_NEI_otherQs}
\end{center}
\vskip -0.2in
\end{figure*}

For logP, SAS and GSK3$\beta$, we see those solutions from BO, are better (on average) than the posterior distribution over AS. The opposite trend is observed for DRD2. For NP score, half of the total $30$ cases ($3$ $Q'$s and $10$ $Q$s) improved from $QoI_\text{post}$, and other halves decreased the $QoI$. The optimization process may explore the subspace regions that are beneficial for the specific $Q$ set but reduce performance (compared to AS posterior) for other design points $Q'$ by moving away from the posterior distribution. This is more pronounced for JNK3. We also see these general trends for the $10$ pairs found using NEI as the acquisition function (\cref{imprv_QoI_NEI_otherQs}). From this exploratory study, it appears that the BO in general changes the tree decoder of JT-VAE focusing specifically on the design points in $Q$. This is not surprising as the optimization did not account for latent space regions beyond $Q$. 

\subsection{Computing Resources}
\label{computing_resources}
Most of the experiments are performed on a workstation with Intel\textsuperscript{\tiny\textregistered} Core i9-11900KF 3.50GHz and single NVIDIA GeForce RTX 3090 GPU. Rest are done using a single node with Intel\textsuperscript{\tiny\textregistered} Xeon 6248R 3.0GHz and single NVIDIA A100 GPU within an HPC cluster.

%%%%%%%%%%%%%%%%%%%%%%%%%%%%%%%%%%%%%%%%%%%%%%%%%%%%%%%%%%%%
\clearpage
\newpage

\end{document}